\documentclass[preprint,12pt]{elsarticle}

\let\today\relax
\makeatletter
\def\ps@pprintTitle{%
    \let\@oddhead\@empty
    \let\@evenhead\@empty
    \def\@oddfoot{\footnotesize\itshape
         {Preprint submitted} \hfill\today}%
    \let\@evenfoot\@oddfoot
    }
\makeatother

\usepackage{hyperref}

\usepackage{amssymb}

\usepackage{lineno}

\usepackage{pgfplots}
\usepackage{pgf, tikz}
 
\usetikzlibrary{arrows, automata}
\usetikzlibrary{shapes.multipart}
\usetikzlibrary{quotes,arrows.meta}%
\usetikzlibrary{decorations.pathreplacing}
\usetikzlibrary{positioning}
\usetikzlibrary{matrix}
\usetikzlibrary{backgrounds}
\usetikzlibrary{fit, patterns}
\usepackage[utf8]{inputenc}
\usepgfplotslibrary{groupplots,dateplot}
\usetikzlibrary{shapes.arrows}
\pgfplotsset{compat=newest}
\usepackage{graphicx}
\usepackage{graphbox}
\usepackage{color}
\usepackage{ifthen}
\usepackage{bm}
\usepackage{xurl}
\usepackage{adjustbox}
\usepackage{subcaption}
\usepackage{comment}
\usepackage{multirow}
\usepackage{xspace}
\usepackage{pgfgantt}

\usepackage{amssymb}
\usepackage{pdfpages}
\makeatletter
\newcommand\saveequation[2]{%
  \@namedef{equation@#1}{#2}%
}
\newcommand\useequation[1]{%
  \@nameuse{equation@#1}%
}
\makeatother
\RequirePackage{amsmath,bm}

\usepackage[figuresright]{rotating}

\usepackage{booktabs}

\usepackage[disable, colorinlistoftodos,textsize=small,textwidth=2.5cm]{todonotes}
\setuptodonotes{backgroundcolor=yellow!10!white}
\newcounter{todoListItems}

\usepackage[normalem]{ulem}

\saveequation{massenergy}{E = mc^{2}}
\saveequation{curlE}{\nabla \times \bm{E} = 0}
\newcommand{\filterNum}[1]{\ensuremath{M^{[#1]}}}
\newcommand{\filterStride}[1]{\ensuremath{s^{[#1]}}}
\newcommand{\filterDilation}[1]{\ensuremath{r^{[#1]}}}
\newcommand{\activationFunction}[1]{\ensuremath{g^{[#1]}}}
\newcommand{\featureMap}[1]{\ensuremath{\bm{A}^{[#1]}}}
\newcommand{\filter}[2]{\ensuremath{F^{[#1]}_{#2}}}
\saveequation{layerActivation}{\featureMap{\ell} = \activationFunction{\ell}(\bm{Z}^{[\ell]}) = \activationFunction{\ell}(\featureMap{\ell-1} * \filter{\ell}{})}

\includecomment{optionalQuestions}
\newcommand{\question}[1]{}
\begin{optionalQuestions}
	\renewcommand{\question}[1]{\noindent\textcolor{gray}{\newline  #1 \newline}}
\end{optionalQuestions}

\newcounter{myexample}[section]

\makeatletter
\DeclareRobustCommand\onedot{\futurelet\@let@token\@onedot}
\def\@onedot{\ifx\@let@token.\else.\null\fi\xspace}

\def\eg{\emph{e.g}\onedot}

\makeatother

\newcommand{\tabhead}[1]{\textbf{#1}}
\providecommand{\decoRule}{\rule{.8\textwidth}{.4pt}} 

\def\minNeuronSize{1.29}%
\def\layerDistance{4.5}
\def\nodeDistance{1.3}
\tikzstyle{neuron} = [rectangle,fill=black!25,minimum size=\minNeuronSize cm,inner sep=0pt]
\tikzstyle{inputNeuronStyle} = [neuron, fill=green!50]
\tikzstyle{outputNeuronStyle}=[neuron, fill=red!50]
\tikzstyle{hiddenNeuronStyle}=[neuron, fill=blue!50]
\tikzstyle{zeroNeuronStyle}=[neuron, fill=gray!20]
\tikzstyle{annotatonStyle}=[text width=3cm,text centered, scale=1]
\tikzstyle{convInfoStyle}=[text width=4cm, text centered, rotate=0, scale=0.9]
\tikzstyle{moduleBox} = [rectangle, draw, inner sep=0pt, dashed, rounded corners,opacity=0.2,fit=#1]
\newcommand{\layerOneD}[3]{%
    \def\layerId{\tempa}
    \def\neuronNum{\tempb}
    \def\zeroPadNumTop{\tempc}
    \def\zeroPadNumBot{\tempd}
    \def\nodeContentM{\tempe} %
    \def\nodeContentSup{\tempf} %
    \def\nodeContentSub{\tempg} %
    \def\stx{\temph}
    \def\sty{\tempi}
    \def\annotationTop{#1}
    \def\annotationBot{#2}
    \def\stl{#3};
    \def\x{\stx}
    \def\y{\sty}
    \ifx\annotationTop\empty
        \pgfmathparse{\y+\nodeDistance}
        \edef\y{\pgfmathresult}
    \else
        \node[annotatonStyle] (\layerId-nt) at (\x, \y) {\annotationTop};
    \fi
    \ifnum\zeroPadNumTop>-1
        \pgfmathparse{\y-\nodeDistance}
        \edef\y{\pgfmathresult}
        \foreach \i in {0,...,\zeroPadNumTop}{
        	\path[xshift=0] 
    	    node[zeroNeuronStyle] (\layerId-zt-\i) at (\x,\y-\nodeDistance*\i) {$0$};
        	}
        \pgfmathparse{\y-\zeroPadNumTop*\nodeDistance}
        \edef\y{\pgfmathresult}
    \fi
    \pgfmathparse{\y-\nodeDistance}
    \edef\y{\pgfmathresult}
    \pgfmathparse{\neuronNum-1}
    \edef\neuronNum{\pgfmathresult}
    \foreach \i in {0,...,\neuronNum}{
    	\path[yshift=0] 
    	node[\stl] (\layerId-\i) at (\x,\y-\nodeDistance*\i) {$\nodeContentM^{\nodeContentSup}_{\i}$};
    }
    \pgfmathparse{\y-\neuronNum*\nodeDistance}
    \edef\y{\pgfmathresult}
    \ifnum \zeroPadNumBot>-1
        \pgfmathparse{\y-\nodeDistance}
        \edef\y{\pgfmathresult}
        \foreach \i in {0,...,\zeroPadNumBot}{
        	\path[yshift=0] 
        	node[zeroNeuronStyle] (\layerId-zb-\i) at (\x,\y-\nodeDistance*\i) {$0$};
        }
       \pgfmathparse{\y-\zeroPadNumBot*\nodeDistance}
        \edef\y{\pgfmathresult}
    \fi
    \ifx\annotationBot\empty
    \else
        \pgfmathparse{\y-\nodeDistance}
        \edef\y{\pgfmathresult}
        \node[annotatonStyle] (\layerId-nb) at (\x, \y) {\annotationBot};
    \fi
}
\newcommand{\dotlayer}[5]{%
    \def\layerId{#1}
    \def\stx{#2}
    \def\sty{#3}
    \def\notationy{#4}
    \def\stl{#5}
    \node[\stl, scale=2] (\layerId-dot) at (\stx, \sty) {$\dots$};
    \node[\stl] (\layerId-dot-n) at (\stx, \notationy) {\dots};
}%
\newcommand{\textOnlyLayer}[4]{%
    \def\layerId{#1}
	\def\stx{#2}
	\def\sty{#3}
	\def\txt{#4}
	\node[convInfoStyle] (\layerId-t) at (\stx, \sty) {\txt};
}%
\newcommand{\layerFlat}[7]{%
	\def\layerId{#1}
	\def\stx{#2}
	\def\sty{#3}
	\def\annotationTop{#4}
	\def\annotationInside{#5} %
	\def\annotationBot{#6}
	\def\stl{#7};
	\def\x{\stx}
	\def\y{\sty}
	\ifx\annotationTop\empty
	\pgfmathparse{\y+\nodeDistance}
	\edef\y{\pgfmathresult}
	\else
	\node[annotatonStyle] (\layerId-nt) at (\x, \y) {\annotationTop};
	\fi
	\pgfmathparse{\y-\nodeDistance}
	\edef\y{\pgfmathresult}
	\node (\layerId) at (\x,\y- 0.5*\minHeight) [\stl] {\rotatebox{90}{\annotationInside}};
	\pgfmathparse{\y-\minHeight}
	\edef\y{\pgfmathresult}
	\ifx\annotationBot\empty
	\else
	\pgfmathparse{\y-\nodeDistance}
	\edef\y{\pgfmathresult}
	\node[annotatonStyle] (\layerId-nb) at (\x, \y) {\annotationBot};
	\fi
}%

 \newcommand{\layerOneDErf}[3]{%
    \def\layerId{\tempa}
    \def\neuronNum{\tempb}
    \def\zeroPadNumTop{\tempc}
    \def\zeroPadNumBot{\tempd}
    \def\nodeContentA{\tempe} %
    \def\nodeContentB{\tempf} %
    \def\nodeContentC{\tempg} %
    \def\stx{\temph}
    \def\sty{\tempi}
    \def\annotationTop{#1}
    \def\annotationBot{#2}
    \def\stl{#3};
    \def\x{\stx}
    \def\y{\sty}
    \ifx\annotationTop\empty
        \pgfmathparse{\y+\nodeDistance}
        \edef\y{\pgfmathresult}
    \else
        \node[annotatonStyle] (\layerId-nt) at (\x, \y) {\annotationTop};
    \fi
    \ifnum\zeroPadNumTop>-1
        \pgfmathparse{\y-\nodeDistance}
        \edef\y{\pgfmathresult}
        \foreach \i in {0,...,\zeroPadNumTop}{
        	\path[xshift=0] 
    	    node[zeroNeuronStyle] (\layerId-zt-\i) at (\x,\y-\nodeDistance*\i) {$-$};
        	}
        \pgfmathparse{\y-\zeroPadNumTop*\nodeDistance}
        \edef\y{\pgfmathresult}
    \fi
    \pgfmathparse{\y-\nodeDistance}
    \edef\y{\pgfmathresult}
    \pgfmathparse{\neuronNum-1}
    \edef\neuronNum{\pgfmathresult}
    \foreach \i in {0,...,\neuronNum}{
        \ifnum \layerId=2
            \def\val{0}
            \ifnum \i=\nodeContentB
                \def\val{1}
            \fi
            \path[yshift=0] node[\stl] (\layerId-\i) at (\x,\y-\nodeDistance*\i) {$\frac{\partial \nodeContentA}{\partial \nodeContentC_{\i}}{=}\val$};
        \else
        	\path[yshift=0] 
        	node[\stl] (\layerId-\i) at (\x,\y-\nodeDistance*\i) {$\frac{\partial \nodeContentA_\nodeContentB}{\partial \nodeContentC_{\i}}$};
        \fi
    }
    \pgfmathparse{\y-\neuronNum*\nodeDistance}
    \edef\y{\pgfmathresult}
    \ifnum \zeroPadNumBot>-1
        \pgfmathparse{\y-\nodeDistance}
        \edef\y{\pgfmathresult}
        \foreach \i in {0,...,\zeroPadNumBot}{
        	\path[yshift=0] 
        	node[zeroNeuronStyle] (\layerId-zb-\i) at (\x,\y-\nodeDistance*\i) {$-$};
        }
       \pgfmathparse{\y-\zeroPadNumBot*\nodeDistance}
        \edef\y{\pgfmathresult}
    \fi
    \ifx\annotationBot\empty
    \else
        \pgfmathparse{\y-\nodeDistance}
        \edef\y{\pgfmathresult}
        \node[annotatonStyle] (\layerId-nb) at (\x, \y) {\annotationBot};
    \fi
}

\makeatletter
\long\def\ifnodedefined#1#2#3{%
    \@ifundefined{pgf@sh@ns@#1}{#3}{#2}%
}
\makeatother

\tikzset{lossLableStyle/.style = {anchor=north west, inner sep=0, outer sep=0,fill opacity=.5,fill=white, text opacity=1, scale=0.5}}
\usepackage{tabularx}
\let\cite\cite

\journal{Image and Vision Computing}

\begin{document}
\newif\ifbone
\bonefalse

\begin{frontmatter}



\title{DDCNet-Multires: Effective Receptive Field Guided Multiresolution CNN for Dense Prediction}


\author{Ali Salehi}
\author{Madhusudhanan Balasubramanian}

\address{Tennessee, United States}

\begin{abstract}
Dense optical flow estimation is challenging when there are large displacements in a scene with heterogeneous motion dynamics, occlusion, and scene homogeneity.  Traditional approaches to handle these challenges include hierarchical and multiresolution processing methods.  Learning-based optical flow methods typically use a multiresolution approach with image warping when a broad range of flow velocities and heterogeneous motion is present.  Accuracy of such coarse-to-fine methods is affected by the \emph{ghosting artifacts} when images are warped across multiple resolutions and by the \emph{vanishing problem} in smaller scene extents with higher motion contrast.  Previously, we devised strategies for building compact dense prediction networks guided by the effective receptive field (ERF) characteristics of the network (DDCNet).  The DDCNet design was intentionally simple and compact allowing it to be used as a building block for designing more complex yet compact networks.  In this work, we extend the DDCNet strategies to handle heterogeneous motion dynamics by cascading DDCNet based sub-nets with decreasing extents of their ERF.  Our DDCNet with multiresolution capability (DDCNet-Multires) is compact without any specialized network layers.  We evaluate the performance of the DDCNet-Multires network using standard optical flow benchmark datasets.  Our experiments demonstrate that DDCNet-Multires improves over the DDCNet-B0 and -B1 and provides optical flow estimates with accuracy comparable to similar lightweight learning-based methods.


\end{abstract}


\end{frontmatter}


\section{Introduction}
\label{sec:introduction}
Having a fast and accurate optical flow estimation method is an essential step in many computer vision applications. Decades of research resulted in many successful dense pixel matching methods. Nevertheless, presence of certain conditions such as large displacements (\eg hundreds of pixels), a high percentage of occlusion, and large homogeneous regions are still challenging.

A powerful machine learning model with a good degree of generalization and easy to fine-tune could be a promising solution for many applications. With learning-based models, many of the assumptions generally employed for modeling and estimating motion from image sequences can be avoided by allowing the models to learn generic transformation of frames in an image sequence.  Deep learning methods have been shown to be successful for dense prediction tasks including dense flow estimation \cite{Long2015, liu2019auto, Lotter2016, zhang2019flow, gur2019single, Eigen2014, ilg2018occlusions}. 

In many computer vision applications, estimating optical flow is an essential intermediate step.  Therefore, compactness, speed, and accuracy of the optical flow estimation algorithms are necessary. Inspired by classical approaches, some deep learning-based flow estimation models have designed custom layers such as explicit feature mapping \cite{dosovitskiy2015flownet, sun2018pwc, hui2018liteflownet}. Those layers proved to be effective in flow estimation. Using standard non-specialized layers in the network, however, can facilitate easier integration of the network within the processing pipeline of other computer vision applications such as activity recognition. 

Estimating large displacements (in the order of hundreds of pixels) in the presence of heterogeneous motion dynamics is one of the main challenges that are yet to be addressed by modern flow estimation algorithms. Multi-resolution approaches have been extensively used in classical and modern methods as a strategy to estimate large flow vectors \cite{anandan1989computational,bergen1992hierarchical, ranjan2017optical}. Many of these methods use spatial pyramid approach to achieve this. In general, a coarser-level optical flow estimated at a given Gaussian pyramid level of frames in an image sequence is used to first warp either a target or reference frame to the next finer level in the pyramidal decomposition.  A finer resolution of optical flow estimate is estimated at the next pyramidal level using the warped image and the corresponding reference or target frames at the respective pyramid level.  This coarse-to-fine level of optical flow estimation process is repeated until the desired or full resolution of optical flow estimates are achieved.  Other hierarchical approaches to estimate large motions in the presence of heterogeneous motion dynamics include using various levels of global flow estimates in a scene or regions within a scene to hierarchically propagate a more accurate initial optical flow estimate to each pixel and thereby allowing the method to converge to the ground truth optical flow quickly.  In this approach, only the flow information is scaled at multiple levels while retaining the original frame resolution \cite{balasubramanian2006computational}.

There are issues associated with methods that use spatial pyramid to deal with large displacements. Inaccurate flow estimates at a coarse level, for example, due to scene occlusion or homogeneity, may propagate to the finer levels. Another important issue with those coarse-to-fine methods is the introduction of \textit{ghosting artifact} during frame warping \cite{janai2018unsupervised}. For example, a foreground object (occluding object) moving over a stationary background will introduce a ghost copy of the occluding object after warping.  Though warping images or feature maps can guide the network to achieve finer-scale estimations and reduce the search range during feature matching, the region with ghosting artifacts will lead to estimation errors due to multiple matching candidate locations.  In addition, smaller objects moving at a higher velocity likely are often ignored (vanished) due to scale mismatch in a coarse-to-fine strategy \cite{ranjan2017optical}. Coarse-to-fine strategies, however, are useful for large motion estimation with magnitudes higher than the extent of the receptive field of deeper neural networks.


Previously, we developed a lightweight deep dilated CNN architecture (DDCNet) and strategies for dense prediction problems \cite{salehi2021ddcnetb0b1}.  In brief, DDCNet-B0 and -B1 were designed based on effective use of dilated convolutional layers for dense prediction tasks. Effective Receptive Field (ERF) of candidate networks was used as guiding principle to design compact networks for dense optical flow estimation. Two key strategies or design recommendations for building DDCNets were: 1) preserving spatial information throughout the network with minimal feature downsampling and 2) designing networks with large enough receptive fields or effective receptive fields.  While techniques such as use of deconvolution layers could be used to recover spatial resolution in a deeper network, preserving resolution of spatial features within the network is essential for dense prediction tasks.  Because the primary task of dense prediction problems such as optical flow estimation is to estimate pixel-level coordinate transformation of a scene, every output unit in the network should have access to large enough spatiotemporal extent of the input image sequences.

In this work, we present a multi-resolution architecture called DDCNet-Multires that neither utilizes image-pyramidal architecture nor any intermediate warping to achieve multi-resolution property. DDCNet-Multires extends the DDCNet design strategies to handle heterogeneous motion dynamics by cascading sub-nets with decreasing extents of their ERF.  With this additional design strategy, coarse optical flow estimates from a sub-net with a larger ERF are refined using one or more sub-nets with narrower ERF.  We demonstrate the performance of DDCNet-Multires model using standard optical flow benchmark datasets.


 
\section{Related Work}
\emph{SpyNet} \cite{ranjan2017optical}, \emph{PWC-Net} \cite{sun2018pwc} and all three variants of \emph{LiteFlowNet} \cite{hui2018liteflownet,hui2019lightweight,hui2020liteflownet3} take advantage of classical coarse-to-fine approach in their designs. SpyNet builds image pyramid and warps the images using estimated flow in the coarser levels.  PWC-Net and LiteFlownets build pyramids from feature maps and do warping on feature maps instead of images.

\emph{FlowNet-Simple} and \emph{FlowNet-Correlation} also achieve some kind of build-in multi-resolution by estimating flow from low-resolution features maps and repeating the process in all consecutive layers of its \emph{refinement} section \cite{dosovitskiy2015flownet}. \emph{FlowNet2} stacks several versions of FlowNets but all its networks work on full-resolution data. It takes advantage of the warping layer to help the higher-level networks to deal with shorter displacements \cite{hui2018liteflownet, hui2020liteflownet3}. Our method differs from FlowNet-Simple: we estimate flow after each sub-net comprised of about 15 layers and each sub-net takes full image features as input. Unlike FlowNet2 we do not use any warping layers.

 \emph{LiteFlowNet2} and \emph{LiteFlowNet3} are designed to address some of the issues associated with propagation of wrong estimated form coarse levels. LiteFlowNet2 increases the speed of the original model along with improvements in the quality of the flow estimates \cite{hui2019lightweight}. LiteFlowNet3 incorporates two additional modifications on LiteFlowNet2 to deal with propagation of wrong flow estimates from coarser levels when there is occlusion or homogeneous regions in a scene \cite{hui2020liteflownet3}.

Effects of the vanishing problem can be reduced by building the pyramid using dense features instead of raw images such as the methods used in PWC-Net and LiteFlowNet \cite{sun2018pwc, hui2020liteflownet3}. The vanishing problem cannot be alleviated even by feature warping technique used in LiteFlowNet \cite{hui2018liteflownet}.  Therefore, a more complicated modification to the cost function was necessary for LiteFlowNet3 to deal with propagation of errors from coarse level to fine levels due to the ghosting problem \cite{hui2020liteflownet3}. DDCNet-Multires integrates DDCNet and multi-resolution strategies in such a way to avoid vanishing and ghosting artifact problems.

DDCNet-B0 shows the efficacy of our compact dense-prediction design strategy  \cite{salehi2021ddcnetb0b1}. Our network architecture is simple while performing better than similar networks such as FlowNet-Simple in terms of speed and accuracy. DDCNet-B1 demonstrated improved effectiveness of the network by dropping odd layers (layers with odd dilation rates) and reducing the resolution of the feature maps. The central part of both of these networks is a sub-net called \emph{flow feature extractor} that consists of several convolution layers with increasing dilation rates (see Figure~\ref{fig:DDCNetParts}). In this work, we incorporate classical multiresolution strategies along with the DDCNet design strategies to further improve the accuracy of dense flow estimation.
    
\section{DDCNet-Multires}
\label{sec:DDCNet}
DDCNet-B0 and DDCNet-B1 networks were designed based on the unique characteristics of dense estimation tasks such as dense flow estimation. Their design is intentionally simple and compact which makes them ideally suited for serving as building blocks for many other applications as well as for building more complex architectures using simpler building blocks. To improve accuracy, we utilize the sub-modules within these basic networks and build more elaborate networks yet within the class of lightweight networks with few trainable parameters and lower processing time.

\begin{sidewaysfigure}[htbp]
    \centering
    \scalebox{0.7}{\def\minWidth{0.32}%
\def\minHeight{5}%
\def\layerDistance{2.2}
\def\nodeDistance{1.3}
           
\trimbox{0.5cm 1cm 1cm 0cm}{
\begin{tikzpicture}[
     shorten >=1pt,->,
     draw=black!50,
     node distance=\layerDistance,
     every pin edge/.style={<-,shorten <=1pt},
     ]
     \tikzstyle{neuron} = [rectangle,fill=black!25,minimum width=\minWidth cm, minimum height=\minHeight cm, inner sep=0pt, anchor=center, font=\footnotesize]
    \tikzstyle{inputNeuronStyle} = [neuron, fill=green!50]
    \tikzstyle{outputNeuronStyle}=[neuron, fill=red!50]
    \tikzstyle{hiddenNeuronStyle}=[neuron, fill=blue!50]
    \tikzstyle{zeroNeuronStyle}=[neuron, fill=gray!20]
    \tikzstyle{annotatonStyle}=[text width=3cm,text centered, scale=1, rotate=90]
    \tikzstyle{convInfoStyle}=[text width=4cm, text centered, rotate=0, scale=0.9]
    \tikzstyle{dotLayerStyle}=[text width=3cm,text centered, scale=0.5]
    \def\hw{\ensuremath{h, w}}
    \def\fNum{64}
    \def\startx{-\minWidth}
    \def\starty{0}
    \layerFlat{f-1}{\startx}{\starty}{$(\hw, 3)$}{Frame 1}{}{inputNeuronStyle}
    
    \def\preprocessLayerNum{3}
    \foreach \id in {1,...,\preprocessLayerNum}{
        \pgfmathparse{\layerDistance*\id}
        \edef\tmpx{\pgfmathresult}
        \layerFlat{p-1-\id}{\tmpx}{\starty}{$(\hw, \fNum)$}{Feature map \id}{}{hiddenNeuronStyle}
    }
    
    \def\starty{-1.5*\minHeight}
    \layerFlat{f-2}{\startx}{\starty}{}{Frame 2}{$(\hw, 3)$}{inputNeuronStyle}
    
    \foreach \id in {1,...,\preprocessLayerNum}{
        \pgfmathparse{\layerDistance*\id}
        \edef\tmpx{\pgfmathresult}
        \layerFlat{p-2-\id}{\tmpx}{\starty}{}{Feature map \id}{$(\hw, \fNum)$}{hiddenNeuronStyle}
    }
    
    \def\starty{-1.5*\minHeight}
    \foreach \id in {1,...,\preprocessLayerNum}{
        \pgfmathparse{\layerDistance*\id-0.5*\layerDistance}
        \xdef\startx{\pgfmathresult}
        \textOnlyLayer{\id}{\startx}{\starty+0.5}{Shared \\ Conv \id \\ \filterNum{\id}=\fNum \\ \filterDilation{\id}=[\id,\id]\\ \activationFunction{\id}=ReLU}
    }
    
     \def\starty{-0.9*\minHeight}
    \pgfmathparse{\startx+0.75*\layerDistance}
    \xdef\startx{\pgfmathresult}
    \textOnlyLayer{c}{\startx}{\starty-0.5*\minHeight}{Concat}
    \pgfmathparse{\startx+0.5*\layerDistance}
    \xdef\startx{\pgfmathresult}
    \pgfmathparse{int(2*\fNum)}
    \edef\chl{\pgfmathresult}
    \layerFlat{conc}{\startx}{\starty}{}{Concatenated maps}{$(\hw, \chl)$}{hiddenNeuronStyle}
    
    \pgfmathparse{\startx+0.5*\layerDistance}
    \xdef\startx{\pgfmathresult}
    
    \def\fNum{128}
    \def\p{1}
    \def\st{[1, 1]}
    \def\hw{{h, w}}
    \foreach \id/\dilation in {4/1, 5/2, 6/4, 7/6, 8/8}{
            \pgfmathparse{\startx+0.5*\layerDistance}
            \xdef\startx{\pgfmathresult}
            \textOnlyLayer{\id}{\startx}{\starty-0.5*\minHeight}{Conv \id \\ \filterNum{\id}=\fNum \\ \filterDilation{\id}=[\dilation,\dilation]\\
            \filterStride{\id}=\st \\\activationFunction{\id}=ReLU}
            \tikzstyle{hiddenNeuronStyle}=[neuron, fill=blue!50,  minimum height=\p*\minHeight cm]
            \pgfmathparse{\startx+0.5*\layerDistance}
            \xdef\startx{\pgfmathresult}
            \layerFlat{\id}{\startx}{\starty}{}{F map \id}{$(\hw, \fNum)$}{hiddenNeuronStyle}
        }
    
    \pgfmathparse{\startx + 0.4*\layerDistance}
    \edef\startx{\pgfmathresult}
    \dotlayer{dot}{\startx}{\starty - 0.5*\minHeight}{\starty- 1.25*\minHeight}{dotLayerStyle}
    
    \pgfmathparse{\startx - 0.1*\layerDistance}
    \edef\startx{\pgfmathresult}
    \foreach \id/\dilation in {18/28, 19/30}{
            \ifnum \id=19
                \pgfmathparse{\startx+0.5*\layerDistance}
                \xdef\startx{\pgfmathresult}
                \textOnlyLayer{\id}{\startx}{\starty-0.5*\minHeight}{Conv \id \\ \filterNum{\id}=\fNum \\ \filterDilation{\id}=[\dilation,\dilation]\\
            \filterStride{\id}=\st \\\activationFunction{\id}=ReLU}
            \fi
            \tikzstyle{hiddenNeuronStyle}=[neuron, fill=blue!50,  minimum height=\p*\minHeight cm]
            \pgfmathparse{\startx+0.5*\layerDistance}
            \xdef\startx{\pgfmathresult}
            \layerFlat{\id}{\startx}{\starty}{}{F map \id}{$(\hw, \fNum)$}{hiddenNeuronStyle}
        }

    \def\fNum{2}
    \def\id{20}
    \pgfmathparse{\startx+1*\layerDistance}
    \xdef\startx{\pgfmathresult}
    \textOnlyLayer{\id}{\startx}{\starty - 0.5*\minHeight}{Conv \id \\ \filterNum{\id}=\fNum \\ \filterDilation{\id}=[1,1] \\ \activationFunction{\id}=Linear }
    \pgfmathparse{\startx+0.5*\layerDistance}
    \xdef\startx{\pgfmathresult}
    \tikzstyle{outputNeuronStyle}=[neuron, fill=red!50,  minimum height=\p*\minHeight cm]
    \layerFlat{\id}{\startx}{\starty}{}{Flow}{$(\hw, \fNum)$}{outputNeuronStyle}

    \draw [line width=1,->, shorten >=0.5cm, shorten <=0.5cm] (f-1) edge (p-1-1);
    \draw [line width=1,->, shorten >=0.5cm, shorten <=0.5cm] (f-2) edge (p-2-1);
     \foreach \i in {1, 2}{
         \pgfmathparse{int(\i + 1)}
         \edef\j{\pgfmathresult}
         \draw [line width=1,->, shorten >=0.5cm, shorten <=0.5cm] (p-1-\i) edge (p-1-\j);
          \draw [line width=1,->, shorten >=0.5cm, shorten <=0.5cm] (p-2-\i) edge (p-2-\j);
      }
    \begin{pgfonlayer}{background}
        
    \coordinate (A) at (0.2*\nodeDistance, -\nodeDistance);
    \coordinate (B) at (3.84*\layerDistance, -9.61*\nodeDistance);
    
    \coordinate (C) at (4.34*\layerDistance, -9.61*\nodeDistance);
    \coordinate (D) at (11.2*\layerDistance, -\nodeDistance);
    \node[moduleBox={(A) (B)}, label={[label distance=1.7*\nodeDistance cm, color=green!20!black]90:Spatial Feature Extractor}, pattern=north west lines, pattern color=green] {}; 
    \node[moduleBox={(C) (D)}, label={[label distance=1.7*\nodeDistance cm, color=blue!20!black]90:Flow Feature Extractor / Flow Feature Refiner}, pattern=north east lines, pattern color=blue] {}; 
    \end{pgfonlayer}
    
\end{tikzpicture}
}}
    \decoRule
    \caption[Building blocks of DDCNet-Multires]{Main components of DDCNet architecture: \emph{spatial feature extractor}) three (or more) convolution layers with dilatation rates of 1, 2, and 3 respectively. \emph{flow feature extractor}) is made of consecutive convolution layers with increasing dilation rates. When dilation rates of this sub-net is constant or only increasing in steps of one, we call this \emph{flow feature refiner}.}
    \label{fig:DDCNetParts}
\end{sidewaysfigure}
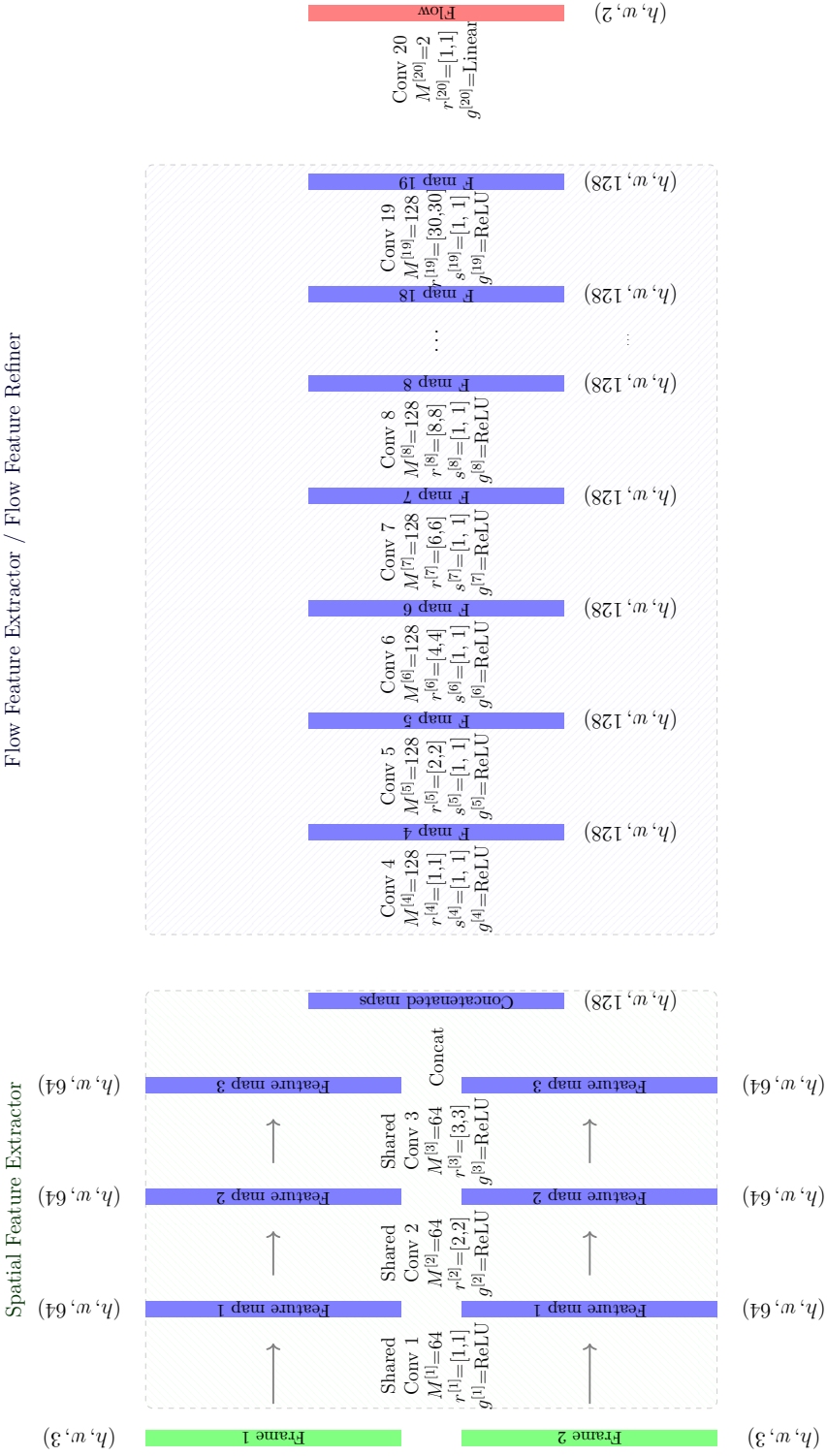

Figure~\ref{fig:DDCNetParts}, shows all important building elements, namely \emph{spatial feature extractor} and \emph{flow feature extractor} that were used for building a multiresolution network called \emph{DDCNet-Multires}. DDCNet-Multires has DDCNet-B1 \cite{salehi2021ddcnetb0b1} as foundation and has all the necessary layers to learn optical flow estimation. In fact, by training this simple network in Figure~\ref{fig:DDCNetParts}, we were able to attain more comparable optical flow estimates with other complicated methods in literature such as FlowNet-Simple.

\emph{Spatial feature extractor} sub-net is comprised of three (or more) convolution layers with dilatation rates of 1, 2, and 3 respectively. It takes the first and second frames of a sequence separately and after applying these shared convolutions on them, concatenates the feature maps from each image and passes them to subsequent layers in the network. \emph{Flow feature extractor} is a sub-net made of consecutive convolution layers with increasing dilation rates. When dilation rates of this sub-net are constant or only increasing in steps of one, we call that \emph{flow feature refiner} (Figure~\ref{fig:DDCNetParts}). Systematic and strategic arrangement of \emph{flow feature extractor} and \emph{flow feature refiner} can be used to build networks with larger receptive fields with appropriate field shape and field smoothness.

\subsection{Multi-resolution}
We utilize our \emph{flow feature extractor} module as a building block for an improved multi-resolution flow estimation network. By adjusting the dilation rates or by passing a downsampled inputs to this module, the ERF of this module can be adjusted to generate intermediate flow estimates at the desired scale or resolution.  For example, using a \emph{flow feature extractor} module with a larger ERF, coarse and large flows in the scene can be estimated.  Therefore, by cascading several \emph{flow feature extractor} modules with decreasing extents of ERFs, initial coarse and large motion estimates can be refined successively to generate finer flow estimates for all flow magnitudes.  

\begin{sidewaysfigure}[htbp]
    \centering
    \scalebox{0.8}{\input{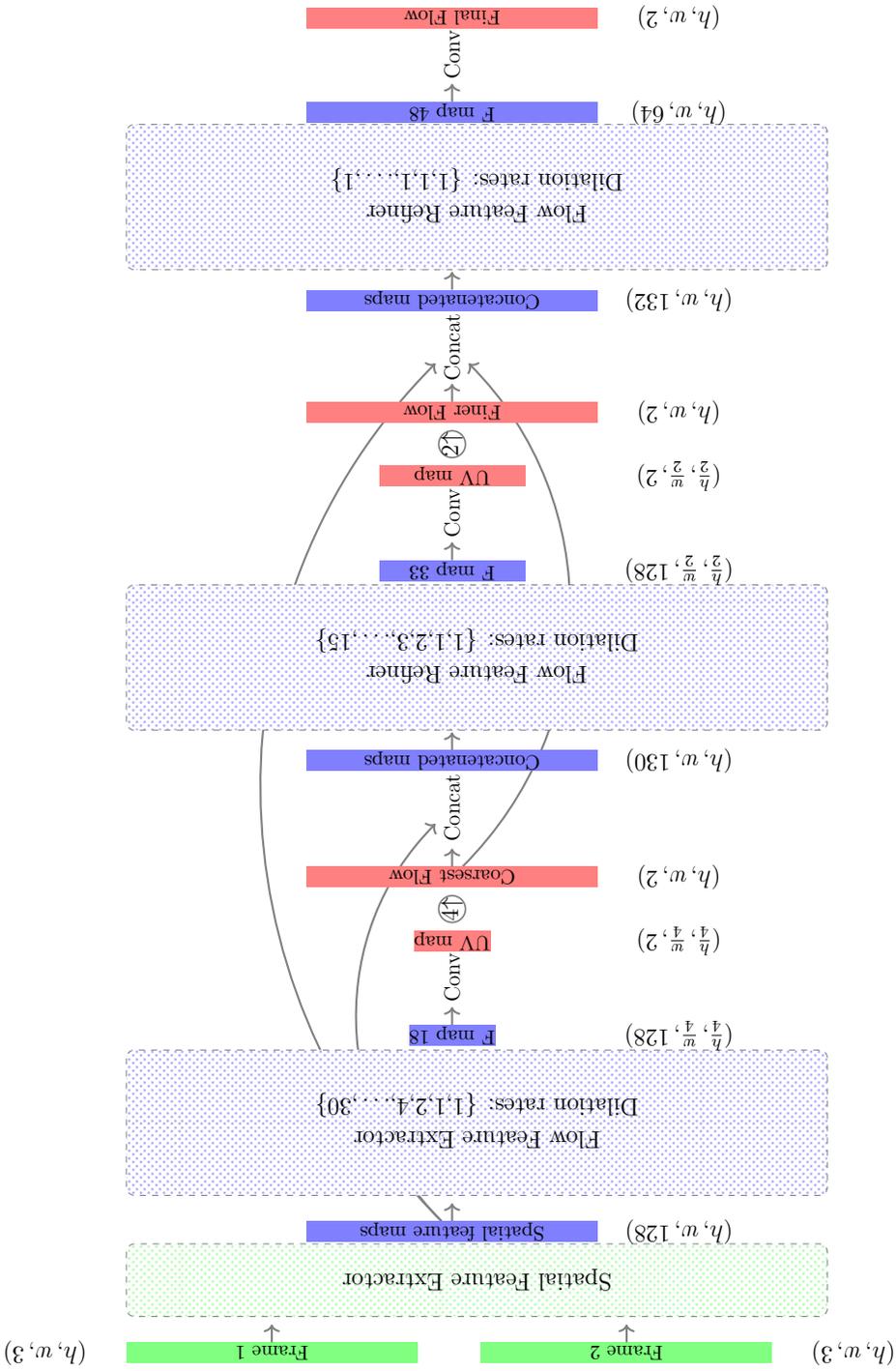}}
    \decoRule
    \caption[DDCNet-Multires]{DDCNet-Multires: A multi-resolution network architecture built using one \emph{spatial feature extractor} module, one \emph{flow feature extractor} module, and two cascaded \emph{flow feature refiner} modules.}
    \label{fig:DDCNetMultires}
\end{sidewaysfigure}

Following the original DDCNet design strategies \cite{salehi2021ddcnetb0b1}, we make sure that the ERF of the DDCNet-Multires architecture covers the majority of large flow vectors i.e. large enough ERF extent, and has suitable ERF shape and smoothness (see ERF of the network in Figure~\ref{fig:ErfMultiresCoarseToFine}). Figure~\ref{fig:DDCNetMultires} shows a multi-resolution network architecture built using one \emph{spatial feature extractor} module, one \emph{flow feature extractor} module, and two cascaded \emph{flow feature refiner} modules. For preprocessing and illumination invariant spatial feature extraction, raw image sequences are fed to a \emph{spatial feature extractor} module resulting in 64 feature maps for each of the images in the sequence. Concatenated spatial feature maps are then passed to a \emph{flow feature extractor} module whose detailed architecture is shown in Figure~\ref{fig:DDCNetParts}. Two convolution filters were applied to the 128 feature maps at 1/4th resolution from the \emph{flow feature extractor} module to generate coarse flow estimates at 1/4th resolution. 

Coarse flow estimates at 1/4th resolution from the \emph{flow feature extractor} were upsampled to full resolution and concatenated with the original spatial feature maps. The concatenated coarse flow estimates and spatial feature maps were fed to a \emph{flow feature refiner} module. Again, two convolutional filters were applied to the 128 feature maps at 1/2 resolution from the first \emph{flow feature refiner} module to generate finer flow estimates at 1/2 resolution. 

Upsampled coarse flow estimates from the \emph{flow feature extractor} module, upsampled finer flow estimates from the first \emph{flow feature refiner} module and the original spatial feature maps were concatenated and fed to a second \emph{flow feature refiner} module comprised of 15 convolutions with no dilation. The final optical flow estimates at the original resolution were obtained by applying one layer of convolution on the feature maps from the second \emph{flow feature refiner} module.

One important distinction with our multi-resolution architecture is that neither do we utilize image-pyramidal architecture nor any intermediate warping to achieve multi-resolution property. Despite naming the different parts of the network as separate modules, they are all sub-nets of a single network.  Therefore, these sub-nets could be more efficiently trained end-to-end using the raw image sequences and the corresponding ground-truth optical flow for each of the training sequences.

\section{Experiments}
\label{sec:experiments}
    \subsection{Datasets}
        \emph{Flying Chairs} dataset was used both for design purposes (selecting more promising networks for further tests) and initializing all other training for faster convergence. This dataset contains 22,872 image pairs with ground truth flow. Sequences were generated by applying random 2D affine transformations of a set of renderings of 3D chair models put on Flicker images from different categories as background \cite{dosovitskiy2015flownet}.
        
\begin{figure}[htbp]
            \centering
            \subcaptionbox{Sintel \label{fig:Histograms:a}}{\scalebox{0.52}{\begin{tikzpicture}
\begin{axis}[
colorbar,
colorbar style={ytick={1,2,3,4,5,6,7,8},yticklabels={\(\displaystyle {10^{1}}\),\(\displaystyle {10^{2}}\),\(\displaystyle {10^{3}}\),\(\displaystyle {10^{4}}\),\(\displaystyle {10^{5}}\),\(\displaystyle {10^{6}}\),\(\displaystyle {10^{7}}\),\(\displaystyle {10^{8}}\)},ylabel={Count}},
colormap/blackwhite,
point meta max=8.15,
point meta min=0.45,
tick align=outside,
tick pos=left,
x grid style={white!69.0196078431373!black},
xlabel={U},
xmin=-199.5, xmax=198.5,
xtick style={color=black},
y grid style={white!69.0196078431373!black},
ylabel={V},
ymin=-199.5, ymax=198.5,
ytick style={color=black}
]
\addplot graphics [includegraphics cmd=\pgfimage,xmin=-199.5, xmax=198.5, ymin=-199.5, ymax=198.5] {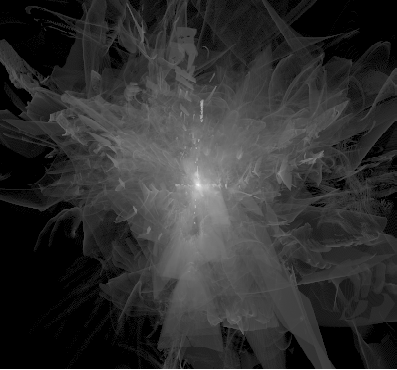};
\end{axis}
\end{tikzpicture}%
}}
            \subcaptionbox{Flying Things3D \label{fig:Histograms:b}}{\scalebox{0.52}{\begin{tikzpicture}

\begin{axis}[
colorbar,
colorbar style={ytick={3,4,5,6,7},yticklabels={\(\displaystyle {10^{3}}\),\(\displaystyle {10^{4}}\),\(\displaystyle {10^{5}}\),\(\displaystyle {10^{6}}\),\(\displaystyle {10^{7}}\)},ylabel={Count}},
colormap/blackwhite,
point meta max=7.92,
point meta min=2.68,
tick align=outside,
tick pos=left,
x grid style={white!69.0196078431373!black},
xlabel={U},
xmin=-199.5, xmax=198.5,
xtick style={color=black},
y grid style={white!69.0196078431373!black},
ylabel={V},
ymin=-199.5, ymax=198.5,
ytick style={color=black}
]
\addplot graphics [includegraphics cmd=\pgfimage,xmin=-199.5, xmax=198.5, ymin=-199.5, ymax=198.5] {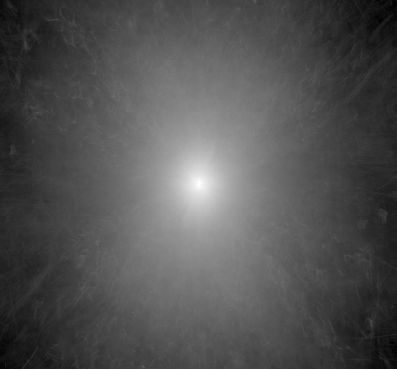};
\end{axis}

\end{tikzpicture}}}
            \decoRule
            \caption[Histogram of flow vectors in Sintel and Flying Things3D datasets]{2D Histogram of flow vectors in Sintel and Flying Things3D datasets with the $u$ and $v$ components of the flow velocities along the horizontal and vertical axes respectively.  Frequency of observing \(u, v\) is in log scale and color coded for clarity.  Each location in the histogram represents the frequency of a specific flow velocity \(u, v\) in the respective dataset with the zero velocity (\((u,v)=[0,0]\)) at the center.}
            \label{fig:Histograms}
        \end{figure}

        \emph{Flying Things3D} is a synthetic stereo flow benchmark dataset with approximately 77 thousand sequences. It provides ground truth stereo disparity and optical flow maps. While the Flying Chairs dataset contains only planar motion, Flying Things3D benchmark sequences have 3D motions and lighting effects and were designed to be more realistic. In addition to linear motions, 3D objects in the scene also undergo non-rigid transformations. Due to large object movements within the scene, optical flow magnitudes, as well as the extent of occlusions, are also larger. Therefore, Flying Things3D dataset, with more diverse scene characteristics and motion dynamics (Figure~\ref{fig:Histograms:b}), is a primary dataset commonly used for training supervised networks for dense flow estimation \cite{mayer2016large}.
        
        \emph{MPI Sintel Dataset} is among the most challenging dataset with higher flow magnitudes, motion heterogeneity, and large occlusions. Two different versions of this dataset namely \emph{clean} and \emph{final} are used in this study. Clean sequences describe specific illumination conditions for the scene. The final sequences include atmospheric effects such as fog along with motion blur. For each dataset version (clean and final), 1,041 sequences with ground truth are available \cite{Butler:ECCV:2012}.
        
        \emph{KITTI2012} and \emph{KITTI2015} are real-word sparse optical flow datasets.  Because the samples are taken from a driver's view, these are limited to only a specific type of motion known as \emph{ego-motion} that captures a rigid scene as the camera movies within the scene. In total, approximately 400 sequences with larger displacements and a larger variety of lighting conditions are present in these datasets. Ground truth occlusion maps are also available \cite{geiger2013vision, menze2015object}.
        
        \emph{Middlebury} includes real scenes with fluorescent tagged textures used for generating ground truth optical flow and realistic computer-generated synthetic scenes. It contains only 8 two-frame sequences with small displacements, around 10 pixels \cite{baker2011database}.
        
    \subsection{Training and Network Details}
        Sub-nets within DDCNet-Multires can be individually trained when there are memory or GPU limitations or the whole DDCNet-Multires with cascaded sub-nets can be trained end-to-end depending on the available GPU capacity.  Initial training for DDCNet-Multires started with the Flying Chair dataset as it is simple in terms of scene dynamics and magnitude of motions. In general, a suitable initial learning rate was identified as the parameter choice resulting in a downward trend in the network error.  Then, we continued training on the Flying Things 3D dataset that has larger motion vectors (see a histogram of its motion vectors in Figure~\ref{fig:Histograms}). This is still a synthetic dataset but with more samples, diverse objects, textures, and lighting conditions. This dataset was designed to have image and scene statistics similar to other benchmark datasets as much as possible.
    
        \emph{Train-time-augmentation} method was used, following the work of \cite{dosovitskiy2015flownet}, to improve accuracy and generalization of the network. Geometric augmentations and photometric augmentations were applied to each batch. For geometric augmentations, all images within each training sequence and the corresponding ground truth optical flow were rotated randomly between -30 and 30 degrees and resized with a random scaling factor uniformly chosen between 0.5 and 1.0. For photometric augmentations, Gaussian noise with zero mean and a standard deviation chosen uniformly from $[0.01, 0.08]$ was added to only images in each sequence. A multiplier for adjusting contrast was chosen randomly from $[0.1, 5]$. Images were converted to HSV and their saturation channels were multiplied by a random saturation factor chosen from $[0.1, 4]$ then converted back to RGB. To modify brightness, a random number, selected from $[-0.3, 0.3]$, was added to all channels of images. After all these perturbations, pixel values were scaled to be in range $[0, 1]$.
        
        The following network training parameters were heuristically obtained.  A batch size of 4 was used for training. Learning rates were set to 0.0001 for the first 900k iterations and divided by 2 after the network loss stays flat for a few epochs (about 60k iterations). As we mentioned in Figure~\ref{fig:DDCNetParts}, all the filters were \(3\times3\) in size and were initialized by the He-initialization method. ReLU activation was used for most of the layers except those that generate flow maps. For the layers that generate flow maps, linear activation was used to generate any real number (positive or negative values).
    
    \subsection{Results}
        \subsubsection{Quantitative and Qualitative Evaluation on Benchmark Datasets}
         Average endpoint error (AEE) metric was used for assessing network performance during our network design and development. We also used $Fl_{\text{all}}$ metric on KITTI datasets, where $Fl_{\text{all}}$ measures the percentage of outlying estimates based on the number of pixel coordinates with an endpoint error (EE) $\ge 3$ pixels of flow and EE $\ge 5\%$ of the magnitude of its ground-truth flow vector were counted as outliers).
         
         \begin{table}[h]
           \centering
           \caption[Quantitative results of different methods on benchmark datasets]{Average endpoint error of selected deep learning-based methods on benchmark datasets. Entries with parentheses indicate the testing performance on data that were previous used for fine-tuning the network. $Fl_{\text{all}}$ measures the percentage of outlier estimates; pixels with EE $\ge 3$  and EE $\ge 5\%$ of the magnitude of its ground-truth flow vector were counted as outliers.
           } 
           \scalebox{0.8}{\scalebox{0.70}{
\begin{tabular}{|c|l||c c|c c|c c|c c c|c c|}
            \hline
\multirow{1}{*}{} 
&\multirow{1}{*}{\tabhead{Method}}   	                             	
&\multicolumn{2}{c|}{\tabhead{Sintel clean}}				
&\multicolumn{2}{c||}{\tabhead{Sintel final}}						
&\multicolumn{2}{c|}{\tabhead{KITTI12}}
&\multicolumn{3}{c||}{\tabhead{KITTI15}}   
&\multicolumn{2}{c|}{\tabhead{Middlebury}}\\

\multirow{1}{*}{}
&\multirow{1}{*}{}
&\multicolumn{1}{c}{train}&\multicolumn{1}{c|}{test}
&\multicolumn{1}{c}{train}&\multicolumn{1}{c||}{test}
&\multicolumn{1}{c}{train}&\multicolumn{1}{c|}{test}
&\multicolumn{1}{c}{train}&\multicolumn{1}{c}{train}&\multicolumn{1}{c||}{test}		
&\multicolumn{1}{c}{train}&\multicolumn{1}{c|}{test}\\	

\multirow{1}{*}{}
&\multirow{1}{*}{}
&\multicolumn{1}{c}{}&\multicolumn{1}{c|}{}
&\multicolumn{1}{c}{}&\multicolumn{1}{c||}{}
&\multicolumn{1}{c}{}&\multicolumn{1}{c|}{}
&\multicolumn{1}{c}{}&\multicolumn{1}{c}{(Fl-all)}&\multicolumn{1}{c||}{ (Fl-all)}		
&\multicolumn{1}{c}{}&\multicolumn{1}{c|}{}\\

\hline\hline    
\multirow{7}{*}{\rotatebox[origin=c]{90}{\tabhead{Heavyweight CNN}}}
&\multirow{1}{*}{FlowNetS~\cite{dosovitskiy2015flownet}}				
&4.50&\multicolumn{1}{c|}{7.42}	           
&5.45&\multicolumn{1}{c||}{8.43}
&8.26&\multicolumn{1}{c|}{}
& &\multicolumn{1}{c}{}&\multicolumn{1}{c||}{}			
&1.09&\multicolumn{1}{c|}{}\\       

\multirow{1}{*}{}
&\multirow{1}{*}{FlowNetS ft-sintel~\cite{dosovitskiy2015flownet}}				
&(3.66)&\multicolumn{1}{c|}{6.96}	           
&(4.44)&\multicolumn{1}{c||}{7.76}
&7.52&\multicolumn{1}{c|}{9.1}
& &\multicolumn{1}{c}{}&\multicolumn{1}{c||}{}			
&0.98&\multicolumn{1}{c|}{} \\ 
                                 
\multirow{1}{*}{}
&\multirow{1}{*}{FlowNetC~\cite{dosovitskiy2015flownet}}				
&4.31&\multicolumn{1}{c|}{7.28}	           
&5.87&\multicolumn{1}{c||}{8.81}
&9.35&\multicolumn{1}{c|}{}	
& &\multicolumn{1}{c}{}&\multicolumn{1}{c||}{}	
&1.15&\multicolumn{1}{c|}{}\\

\multirow{1}{*}{}
&\multirow{1}{*}{FlowNetC ft-sintel~\cite{dosovitskiy2015flownet}}				
&(3.78)&\multicolumn{1}{c|}{6.85}	           
&(5.28)&\multicolumn{1}{c||}{8.51}
&8.79&\multicolumn{1}{c|}{}	
& &\multicolumn{1}{c}{}&\multicolumn{1}{c||}{}		
&0.93&\multicolumn{1}{c|}{} \\
  



\multirow{1}{*}{}
&\multirow{1}{*}{FlowNet2~ \cite{ilg2017flownet}}				
&{2.02}&\multicolumn{1}{c|}{{3.96}}	           
&{3.54}&\multicolumn{1}{c||}{6.02}
&4.01&\multicolumn{1}{c|}{}
&10.08&\multicolumn{1}{c}{29.99\%}&\multicolumn{1}{c||}{}			
&{0.35}&\multicolumn{1}{c|}{{0.52}} \\ 

\multirow{1}{*}{}                                                                                            
&\multirow{1}{*}{FlowNet2 ft-sintel~\cite{ilg2017flownet}}				
&(1.45)&\multicolumn{1}{c|}{4.16}	           
&(2.19)&\multicolumn{1}{c||}{{5.74}}
&{3.54}&\multicolumn{1}{c|}{}
&{9.94}&\multicolumn{1}{c}{{28.02\%}}&\multicolumn{1}{c||}{}			
&{0.35}&\multicolumn{1}{c|}{} \\ 

\multirow{1}{*}{}
&\multirow{1}{*}{FlowNet2 ft-kitti~\cite{ilg2017flownet}}				
&3.43&\multicolumn{1}{c|}{}	           
&4.83&\multicolumn{1}{c||}{}
&(1.43)&\multicolumn{1}{c|}{{1.8}}
&(2.36)&\multicolumn{1}{c}{(8.88\%)}&\multicolumn{1}{c||}{{11.48\%}}			
&0.56&\multicolumn{1}{c|}{} \\
\multirow{1}{*}{}
&\multirow{1}{*}{}				
&&\multicolumn{1}{c|}{}	           
&&\multicolumn{1}{c||}{}
&&\multicolumn{1}{c|}{{}}
&&\multicolumn{1}{c}{}&\multicolumn{1}{c||}{{}}			
&&\multicolumn{1}{c|}{} \\

\hline\hline        
\multirow{12}{*}{\rotatebox[origin=c]{90}{\tabhead{Lightweight CNN}}}
\multirow{1}{*}{}
&\multirow{1}{*}{SPyNet~\cite{ranjan2017optical}}				
&4.12&\multicolumn{1}{c|}{6.69}	           
&5.57&\multicolumn{1}{c||}{8.43}
&9.12&\multicolumn{1}{c|}{}
& &\multicolumn{1}{c}{}&\multicolumn{1}{c||}{}		
&{0.33}&\multicolumn{1}{c|}{{0.58}} \\  

\multirow{1}{*}{}
&\multirow{1}{*}{SPyNet ft-sintel~\cite{ranjan2017optical}}				
&(3.17)&\multicolumn{1}{c|}{6.64}	           
&(4.32)&\multicolumn{1}{c||}{8.36}
&{3.36}&\multicolumn{1}{c|}{4.1}	
& &\multicolumn{1}{c}{}&\multicolumn{1}{c||}{35.07\%}	
&0.33&\multicolumn{1}{c|}{0.58} \\  

\multirow{1}{*}{}
&\multirow{1}{*}{PWC-Net+ ft-sintel~\cite{sun2019models}}				
&(1.71)&\multicolumn{1}{c|}{3.45}	           
&(2.34)&\multicolumn{1}{c||}{4.60}
&&\multicolumn{1}{c|}{}	
& &\multicolumn{1}{c}{}&\multicolumn{1}{c||}{}	
&&\multicolumn{1}{c|}{} \\

\multirow{1}{*}{}
&\multirow{1}{*}{PWC-Net+ ft-kitti~\cite{sun2019models}}				
&&\multicolumn{1}{c|}{}	           
&&\multicolumn{1}{c||}{}
&(0.99)&\multicolumn{1}{c|}{1.4}	
&(1.47)&\multicolumn{1}{c}{(7.59\%)}&\multicolumn{1}{c||}{7.72\%}	
& &\multicolumn{1}{c|}{} \\

\multirow{1}{*}{}
&\multirow{1}{*}{LiteFlowNet~\cite{hui2018liteflownet}}				
&{2.48}&\multicolumn{1}{c|}{}	           
&{4.04}&\multicolumn{1}{c||}{}
&{4.00}&\multicolumn{1}{c|}{}		
&{10.39}&\multicolumn{1}{c}{{28.50\%}}&\multicolumn{1}{c||}{}		
&0.39&\multicolumn{1}{c|}{} \\   

\multirow{1}{*}{}
&\multirow{1}{*}{LiteFlowNet ft-sintel~\cite{hui2018liteflownet}}				
&(1.64)&\multicolumn{1}{c|}{{4.86}}	           
&(2.23)&\multicolumn{1}{c||}{6.09}
&&\multicolumn{1}{c|}{}		
&&\multicolumn{1}{c}{}&\multicolumn{1}{c||}{}		
&&\multicolumn{1}{c|}{} \\

\multirow{1}{*}{}
&\multirow{1}{*}{LiteFlowNet ft-kitti~\cite{hui2018liteflownet}}				
&&\multicolumn{1}{c|}{}	           
&&\multicolumn{1}{c||}{}
&(1.29)&\multicolumn{1}{c|}{{1.7}}		
&(2.16)&\multicolumn{1}{c}{(8.16\%)}&\multicolumn{1}{c||}{{10.24\%}}		
&&\multicolumn{1}{c|}{} \\   

\multirow{1}{*}{}
&\multirow{1}{*}{LiteFlowNet2 ft-sintel~\cite{hui2019lightweight}}				
&(1.30)&\multicolumn{1}{c|}{3.48}	           
&(1.62)&\multicolumn{1}{c||}{4.69}
&&\multicolumn{1}{c|}{}	
& &\multicolumn{1}{c}{}&\multicolumn{1}{c||}{}	
&&\multicolumn{1}{c|}{} \\

\multirow{1}{*}{}
&\multirow{1}{*}{LiteFlowNet2 ft-kitti~\cite{hui2019lightweight}}				
&&\multicolumn{1}{c|}{}	           
&&\multicolumn{1}{c||}{}
&(0.95)&\multicolumn{1}{c|}{1.4}	
&(1.33)&\multicolumn{1}{c}{(4.32\%)}&\multicolumn{1}{c||}{7.62\%}	
&&\multicolumn{1}{c|}{} \\

\multirow{1}{*}{}
&\multirow{1}{*}{LiteFlowNet3 ft-sintel~\cite{hui2020liteflownet3}}				
&(1.32)&\multicolumn{1}{c|}{2.99}	           
&(1.76)&\multicolumn{1}{c||}{4.45}
&&\multicolumn{1}{c|}{}	
& &\multicolumn{1}{c}{}&\multicolumn{1}{c||}{}	
&&\multicolumn{1}{c|}{} \\

\multirow{1}{*}{}
&\multirow{1}{*}{LiteFlowNet3 ft-kitti~\cite{hui2020liteflownet3}}				
&&\multicolumn{1}{c|}{}	           
&&\multicolumn{1}{c||}{}
&(0.91)&\multicolumn{1}{c|}{1.3}	
&(1.26)&\multicolumn{1}{c}{(3.82\%)}&\multicolumn{1}{c||}{7.34\%}	
&&\multicolumn{1}{c|}{} \\

\cline {2-13}

\multirow{1}{*}{}
&\multirow{1}{*}{DDCNet-B0  ft-sintel}				
&(2.71)&\multicolumn{1}{c|}{7.20}           
&(3.27)&\multicolumn{1}{c||}{7.46}
&7.35&\multicolumn{1}{c|}{}		
&15.29&\multicolumn{1}{c}{47.78\%}&\multicolumn{1}{c||}{{}}		
&0.67&\multicolumn{1}{c|}{} \\    

\multirow{1}{*}{}
&\multirow{1}{*}{DDCNet-B1}				
&4.12&\multicolumn{1}{c|}{}           
&5.46&\multicolumn{1}{c||}{}
&9.57&\multicolumn{1}{c|}{}		
&16.43&\multicolumn{1}{c}{59.03\%}&\multicolumn{1}{c||}{}
&1.2&\multicolumn{1}{c|}{} \\  

\multirow{1}{*}{}
&\multirow{1}{*}{DDCNet-B1  ft-sintel}				
&(1.96)&\multicolumn{1}{c|}{6.19}           
&(2.25)&\multicolumn{1}{c||}{6.91}
&6.65&\multicolumn{1}{c|}{}		
&13.22&\multicolumn{1}{c}{52.68\%}&\multicolumn{1}{c||}{{}}		
&1.14&\multicolumn{1}{c|}{} \\  

\multirow{1}{*}{}
&\multirow{1}{*}{DDCNet-B1  ft-kitti}				
&6.65&\multicolumn{1}{c|}{}           
&8.38&\multicolumn{1}{c||}{}
&(1.76)&\multicolumn{1}{c|}{4.2}		
&(2.57)&\multicolumn{1}{c}{(15.56)\%}&\multicolumn{1}{c||}{38.23}	
&1.74&\multicolumn{1}{c|}{} \\  

\ifbone
\else
    \multirow{1}{*}{}
    &\multirow{1}{*}{DDCNet-Multires}			
    &2.71&\multicolumn{1}{c|}{}           
    &4.14&\multicolumn{1}{c||}{}
    &5.95&\multicolumn{1}{c|}{}		
    &13.54&\multicolumn{1}{c}{43.12\%}&\multicolumn{1}{c||}{}	
    &0.49&\multicolumn{1}{c|}{} \\
    
    \multirow{1}{*}{}
    &\multirow{1}{*}{DDCNet-Multires ft-sintel}			
    &{(1.36)}&\multicolumn{1}{c|}{5.34}           
    &{(1.70)}&\multicolumn{1}{c||}{{5.86}}
    &5.41&\multicolumn{1}{c|}{}		
    &12.59&\multicolumn{1}{c}{40.30\%}&\multicolumn{1}{c||}{}		
    &0.58&\multicolumn{1}{c|}{} \\
    
    \multirow{1}{*}{}
    &\multirow{1}{*}{DDCNet-Multires ft-kitti}			
    &6.86&\multicolumn{1}{c|}{}           
    &7.54&\multicolumn{1}{c||}{}
    &{(0.92)}&\multicolumn{1}{c|}{{3.2}}		
    &{(1.33)}&\multicolumn{1}{c}{{(5.59\%)}}&\multicolumn{1}{c||}{24.66\%}		
    &0.72&\multicolumn{1}{c|}{} \\
\fi

\hline
\end{tabular}
}}
           \label{tab:QuantitativeBenchmark}
        \end{table}
         
         Quantitative performance of our DDCNet models and other state-of-the-art lightweight and heavyweight deep learning models for the Sintel, KITTI12, KITTI15, and Middlebury datasets are presented in Table~\ref{tab:QuantitativeBenchmark}. For datasets with large displacements such as Sintel, DDCNet-Multires outperformed DDCNet-B0 and DDCNet-B1. This can be observed from sequences with large displacements in them. For example, B0 fails in regions with motions larger than 250 pixels in the `large motion' sequence (second row) in Figure~\ref{fig:QualitativeB0vsB1vsMultires} and in Figure~\ref{fig:QualitativeB0vsB1vsMultiresError} whereas B1 performs better and Multires has the best performance. But for Middlebury with small to moderate flow velocities, B0 performed better than B1. The endpoint error of DDCNet-Multires is less than B0 even on the Middlebury dataset demonstrating the effectiveness of the flow refiner modules.

         One of the principal applications of the DDCNet sub-nets is in its use as a building block for designing lightweight, more efficient, and more effective optical flow estimation networks such as the DDCNet-Multires. FlowNet-Simple has been used as a building block for numerous optical flow estimation methods \cite{ilg2017flownet, jason2016back, ren2017unsupervised, wang2018occlusion}. LiteFlowNet and its variants are among the best performing lightweight learning-based models \cite{hui2018liteflownet, hui2020liteflownet3}.  Therefore, we focus on comparing the performance of our DDCNet-Multires primarily with FlowNet and LiteFlowNet.
          
\begin{sidewaysfigure}
          \centering
          \scalebox{0.9}{\def\scale{0.166}
\def\rootDir{Figures/Qualitative/Sintel}
\newcommand{\rott}[1]{\fontsize{1}{1}\selectfont{\rotatebox[origin=c]{90}{#1}}}

\newcommand{\image}[2]{\begin{tikzpicture}[baseline=(current bounding box.center)]%
    \node[anchor=north west, inner sep=0, outer sep=0] (image) at (0,0){\includegraphics[align=c,width=\scale\textwidth]{\rootDir/#1}};%
    \def\temp{#2}\ifx\temp\empty%
    \else
       \node[lossLableStyle](loss) at (0.1,-0.1) {EPE: #2};%
    \fi%
\end{tikzpicture}}%

\newcommand{\rowImg}[6]{
\xdef\id{#2}
\begin{minipage}[b]{0.01\linewidth}\centering \rott{#1}\end{minipage} & \image{Frames/\id.png}{} & \image{GroundTruth/\id.png}{} & \image{FlowNetS/Flow/\id.png}{#3} & \image{FlowNet2/Flow/\id.png}{#4} & \image{LiteFlowNet3/Flow/\id.png}{#5} & \ifbone \image{DDCNetB1/Flow/\id.png}{#6} \else \image{DDCNetMultires/Flow/\id.png}{#6} \fi
}

\newcommand{\capNew}[1]{\begin{minipage}[b]{\scale\linewidth}\centering\subcaption{\tiny #1}\end{minipage}}
\setlength\tabcolsep{1.5pt}

\begin{tabular}{ccccccc}
    \rowImg{Market 3}{01}{1.66}{1.32}{1.10}{\ifbone 2.28 \else 1.62 \fi}\\
    \rowImg{Shaman 1}{02}{1.09}{0.79}{0.46}{\ifbone 1.65 \else 0.89 \fi}\\
    \rowImg{Ambush 1}{03}{37.30}{37.50}{33.17}{\ifbone 35.50 \else 30.73 \fi}\\
    \rowImg{Ambush 3}{04}{10.47}{6.97}{4.85}{\ifbone 7.60 \else 7.12 \fi}\\
    \rowImg{Bamboo 3}{05}{1.22}{1.02}{0.96}{\ifbone 1.99 \else 1.37 \fi}\\
    \rowImg{Cave 3}{06}{3.87}{5.45}{4.64}{\ifbone 8.27 \else 7.10 \fi}\\
    & \capNew{Reference Frame} & \capNew{Ground Truth} & \capNew{FlowNet Simple} & \capNew{FlowNet2} & \capNew{LiteFlowNet3} & \ifbone \capNew{DDCNet B1}
    \else \capNew{DDCNet Multires} \fi%
\end{tabular}

}
          \decoRule
          \caption[Qualitative comparison between our methods and others others on Sintel]{Performance of our DDCNet-Multires model, LiteFlowNet3 (lightweight), FlowNet-Simple (heavyweight) and FlowNet2 (heavyweight) models for estimating optical flow for selected Sintel image sequences.}%
          \label{fig:QualitativeSintel}%
        \end{sidewaysfigure}
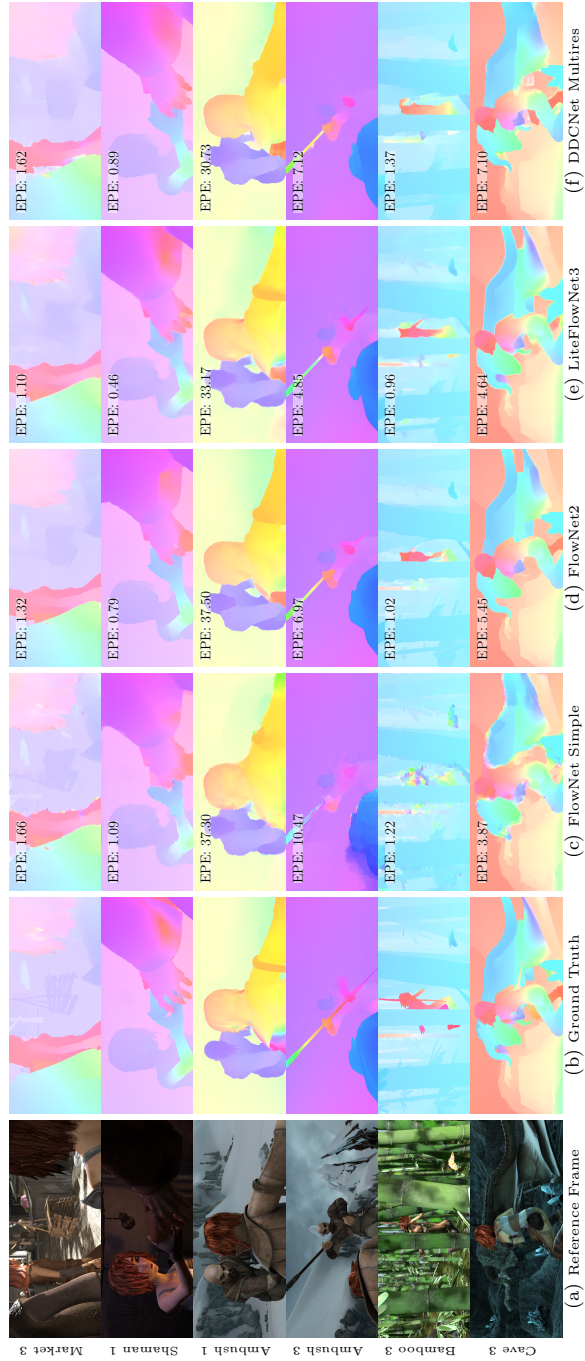

          Figure~\ref{fig:QualitativeSintel} shows the ground truth optical flow and optical flow estimated by FlowNet-Simple, FlowNet2, LiteFlowNet3, and DDCNet-Multires on Sintel dataset.  As summarized in Table~\ref{tab:RunTimeAndModelSize}, FlowNet-Simple has 6 times more parameters (38 million vs 5.54) than DDCNet-Multires. Despite having an additional variational-based refiner step on top of the main network, FlowNet-Simple has a higher endpoint error on almost all of the benchmark datasets. Superior quality of DDCNet-Multires optical flow estimates when compared to FlowNet-Simple can be observed in Figure~\ref{fig:QualitativeSintel}.

        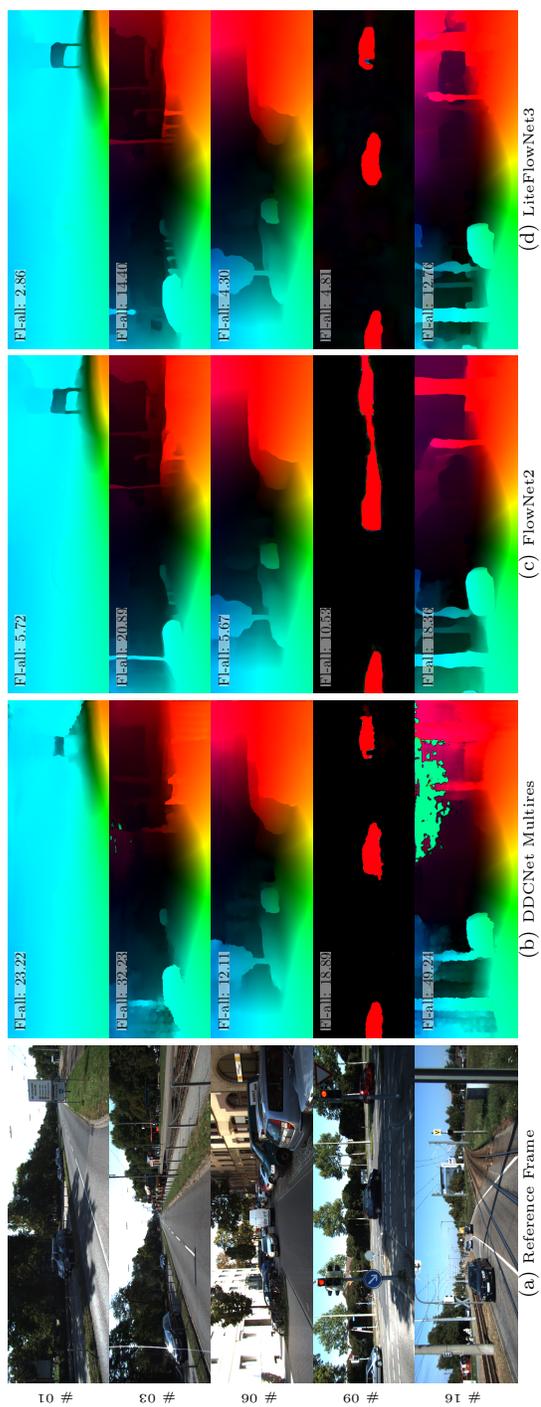
\begin{sidewaysfigure}
          \centering
          \scalebox{0.95}{\def\scale{0.245}
\def\rootDir{Figures/Qualitative/Kitti2015}
\newcommand{\rott}[1]{\fontsize{1}{1}\selectfont{\rotatebox[origin=c]{90}{#1}}}

\newcommand{\image}[2]{\begin{tikzpicture}[baseline=(current bounding box.center)]%
    \node[anchor=north west, inner sep=0, outer sep=0] (image) at (0,0){\includegraphics[align=c,width=\scale\textwidth]{\rootDir/#1}};%
    \def\temp{#2}\ifx\temp\empty%
    \else
       \node[lossLableStyle](loss) at (0.1,-0.1) {Fl-all: #2};%
    \fi%
\end{tikzpicture}}%
    
\newcommand{\rowImg}[4]{
\xdef\id{#1}
\begin{minipage}[b]{0.01\linewidth}\centering \rott{\# #1} \end{minipage} &
\image{Frames/\id.png}{} & \image{DDCNetMultires/\id.png}{#2} & \image{FlowNet2/\id.png}{#3} & \image{LiteFlowNet3/\id.png}{#4}
}

\newcommand{\capNew}[1]{\begin{minipage}[b]{\scale\linewidth}\centering\subcaption{\tiny #1}\end{minipage}}
\setlength\tabcolsep{1.5pt}
\renewcommand{\arraystretch}{0} 
\begin{tabular}{ccccc}
    \rowImg{01}{23.22}{5.72}{2.86}\\
    \rowImg{03}{32.23}{20.89}{14.40}\\
    \rowImg{06}{12.11}{5.67}{4.30}\\
    \rowImg{09}{18.89}{10.53}{4.81}\\
    \rowImg{16}{49.24}{18.36}{12.76}\\
     &\capNew{Reference Frame} & \capNew{DDCNet Multires} & \capNew{FlowNet2} & \capNew{LiteFlowNet3} 
\end{tabular}}
          \decoRule
          \caption[Qualitative comparison between our methods and others on Kitti2015]{Performance of our DDCNet-Multires model, LiteFlowNet3 (lightweight) and FlowNet2(heavyweight) models for estimating optical flow for selected Kitti2015 image sequences.}%
          \label{fig:QualitativeKitti2015}%
        \end{sidewaysfigure}
         
        FlowNet2 is modeled as a stack of several sub-networks. The endpoint error of FlowNet2 is better than our Multires model, but FlowNet2 has 25 times more learnable parameters than the Multires model. Further, FlowNet2 is more than 6 times slower during testing and it is expected to require a longer training duration since each network needs to be trained separately. We could likely improve the performance of DDCNets using a similar stacking strategy while retaining the strengths of our network modeling approach. On Kitti2015 dataset, while FlowNet2 has a lower Fl-all error metric compared to DDCNet-Multires, it fails when it comes to motion boundaries (it can be seen in the result of sample \#09 in Figure~\ref{fig:QualitativeKitti2015}).
        
        Among lightweight methods, DDCNet-Multires outperforms SPyNet on challenging Sintel and Kitti datasets but falls behind PWC-Net. LiteFlowNet models perform better than our DDCNet-Multires on Sintel clean test data. Our model outperforms LiteFlowNet on more challenging Sintel final test dataset. LiteFlowNet2 and LiteFlowNet3 have complicated designs but outperform all other methods on Sintel and Kitti datasets.

        \subsubsection{Model Size / Compactness and Processing Time}
        Table~\ref{tab:RunTimeAndModelSize} shows a summary of model parameters (number of layers, and number of learnable parameters) and computational speed of processing Sintel image sequences. 

        \begin{table}
            \centering
            \caption[Number of trainable parameters and computational of models]{Number of trainable parameters and computational speed of processing Sintel image sequences for lightweight and heavyweight models. Runtime is measured using Sintel image sequences with a frame size of \(1024 \times 436\) pixels. General speed differences between the computational frameworks such as Tensorflow and Caffe should be considered when comparing the run times of various networks.}
            \scalebox{0.6}{\begin{tabular}{ l|l|l|l|l|l|l}
			\toprule
			\tabhead{Method} & \tabhead{Number of} & \tabhead{Number of} &  \tabhead{Framework} & \tabhead{GPU (NVIDIA)} & \tabhead{Time (ms)} & \tabhead{FPS} \\
			 & \tabhead{layers} & \tabhead{parameters (m)} & & & & \\
			\midrule
			\tabhead{DDCNet-B0} & 31 & 1.03 & TF2 &  Quadro RTX 8000 & 76 & 13\\
			\tabhead{DDCNet-B1}  & 30 & 2.99 & TF2 &  Quadro RTX 8000 & 30 & 33\\
			\tabhead{DDCNet-Multires}  & 52 & 5.54 & TF2 &  Quadro RTX 8000 & 88 & 11\\
			  &  &  & \textit{Possible Caffe} &  Quadro RTX 8000 & \textit{17} & \textit{58}\\
			\hline\\
			\tabhead{FlowNet Simple} & 17 & 38 & TF1 &   Tesla K80 & 86 & 11\\
			 &  &  & Caffe &     GTX  1080 & 18 & 55\\
			\tabhead{FlowNet Correlation} & 26 & 39.16 & TF1 &   Tesla K80 & 179 & 5\\
			 &  &  & Caffe &    GTX 1080 & 32 & 31\\
			\tabhead{FlowNet2} & 115 & 162.49 & TF1 &   Tesla K80 & 692 & 1\\ 
			 &  &  & Caffe &    GTX 1080 & 123 & 8\\ 
			\hline\\
			\tabhead{LiteFlowNet} & 94 & 5.37 & Caffe &    GTX  1080 & 88.53 & 12\\
			\tabhead{SPyNet} & 35 & 1.2 & Torch &    GTX  1080 & 129.83 & 8 \\
			\tabhead{PWC-Net+} & 59 & 8.75 & Caffe &  TITAN Xp & 39.63 & 25\\
			\bottomrule
\end{tabular}}
            \label{tab:RunTimeAndModelSize}
        \end{table}
        
        The number of trainable parameters for DDCNet-B0 and DDCNet-B1 was 1.03 and 2.99 million respectively. DDCNet-Multires has 5.54 million parameters which is comparable with the current compact / lightweight optical flow models.  In contrast, more elaborate and heavyweight models such as FlowNet models have 38 million to 162 million trainable parameters.
        
        All of the DDCNets were developed using Tensorflow whereas other lightweight models were mainly implemented using Caffe. FlowNet implementations are available in both Tensorflow and Caffe. In general, Caffe implementations are several times faster than Tensorflow models. Since our GPU has more RAM, we limited the batch size to 1 to make test times comparable. We report the average time for running one thousand batches of image sequences.  With 30 ms processing time in Tensorflow, DDCNet-B1 model is about 3 times faster than Tensorflow implementation of FlowNet-Simple on a less powerful GPU. B1 is also 3 times faster than LiteFlowNet implemented using Caffe and runs on GTX 1080. DDCNet-Multires has test time comparable to FlowNet-Simple and accuracy close to FlowNet2 while being several times faster than it.
        
        \begin{figure*}[htbp]
            \centering
            \scalebox{0.8}{\input{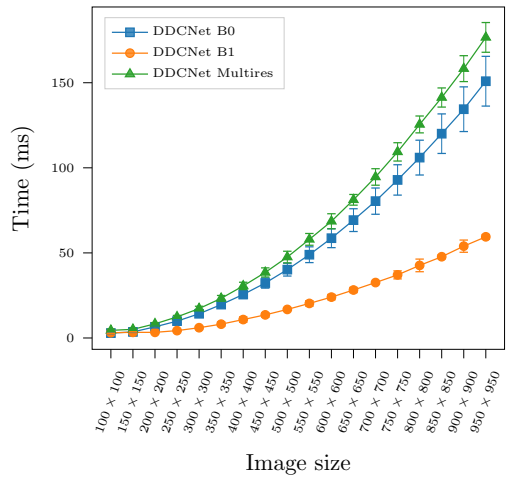}}
            \decoRule
            \caption[Testing time of the DDCNet models as a function of sequence frame size]{Testing time of the DDCNet models as a function of sequence frame size.  B1 is computationally more efficient than B0 while being more accurate. Tests are performed on a single GPU Quadro RTX 8000 with batch sizes of 1.}
            \label{fig:TimeB0vsB1}
        \end{figure*}
        
        Figure~\ref{fig:TimeB0vsB1} shows the frame processing rate of DDCNet models as a function of frame sizes. All DDCNet models are very fast (processing time per frame \(< 10\) ms) for frame size smaller than \(200 \times 200 \) pixels. B1 is the fastest model across all frame sizes. Though B0 has the least number of learnable parameters with full resolution feature maps within the network, the number of floating-point operations is much higher compared to the B1 model.
    
    \subsection{Ablation study}
    In order to understand the efficacy of the designed DDCNet-Multires, in this section, we will compare its results on several challenging test cases from Sintel dataset. Later we will investigate its receptive field and intermediate flow estimates to see effectiveness of its sub-nets.
    
    
    \subsection{Visual Inspection of Flow Estimates by Different DDCNet Models}
\begin{sidewaysfigure}
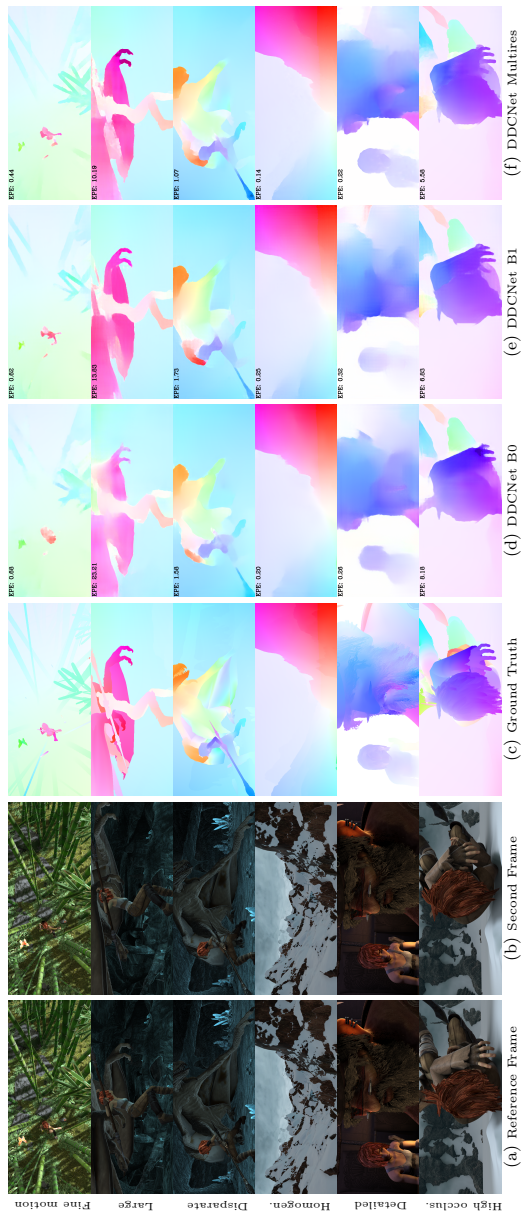

      \centering
      \scalebox{0.8}{\def\scale{0.166}
\def\rootDir{Figures/Qualitative/DDCNets}
\newcommand{\rott}[1]{\fontsize{1}{1}\selectfont{\rotatebox[origin=c]{90}{#1}}}

\newcommand{\IMG}[1]{\includegraphics[align=c,width=\scale\textwidth]{\rootDir/#1}}
\newcommand{\rowImg}[2]{
\xdef\id{#2}
\begin{minipage}[b]{0.01\linewidth}\centering \rott{#1} \end{minipage} & \IMG{GroundTruth/\id-f.png} & \IMG{GroundTruth/\id-s.png} & \IMG{GroundTruth/\id-gt.png} & \IMG{B0/\id-pr.png} & \IMG{B1/\id-pr.png} & \IMG{Multires/\id-pr.png}
}

\newcommand{\capNew}[1]{\begin{minipage}[b]{\scale\linewidth}\centering\subcaption{\tiny #1}\end{minipage}}
\setlength\tabcolsep{1.5pt}
\begin{tabular}{ccccccc}
    \rowImg{Fine motion}{152}\\
    \rowImg{Large}{243}\\
    \rowImg{Disparate}{270}\\
    \rowImg{Homogen.}{364}\\
    \rowImg{Detailed}{391}\\
    \rowImg{High occlus.}{64}\\
    & \capNew{Reference Frame} & \capNew{Second Frame} & \capNew{Ground Truth} & \capNew{DDCNet B0} & \capNew{DDCNet B1} & \capNew{DDCNet Multires} 
\end{tabular}
}
      \decoRule
      \caption[Qualitative results: B0 vs B1 vs Multires]{Optical flow estimates of DDCNet-B0, B1 vs Multires for qualitative assessment of the network performance.  Corresponding error maps are shown in \ref{fig:QualitativeB0vsB1vsMultiresError}}%
      \label{fig:QualitativeB0vsB1vsMultires}%
    \end{sidewaysfigure}

    Figure~\ref{fig:QualitativeB0vsB1vsMultires} shows performance of each of the DDCNet models on a broad variety of six example sequences with varied scene composition and motion dynamics from the \emph{Sintel Training Clean datasets}. The examples are namely 1) \emph{fine motion} sequence (first row in Figure~\ref{fig:QualitativeB0vsB1vsMultires}) involving finer motions of smaller objects in various directions; 2) \emph{large motion} sequence (second row) involving higher flow velocities; 3) \emph{disparate motion} sequence (third row) involving varied flow dynamics in various parts of the scene with varied flow velocities; 4) \emph{homogeneous texture} sequence (fourth row) with motions involving scene segments with homogeneous texture; 5) \emph{high texture} sequence (fifth row) with motions involving highly textured scene segments; and 6) \emph{high occlusion} sequence (sixth row) with aperture problem, entire or partial object views entering the scene and / or leaving the scene.  

\begin{sidewaysfigure}
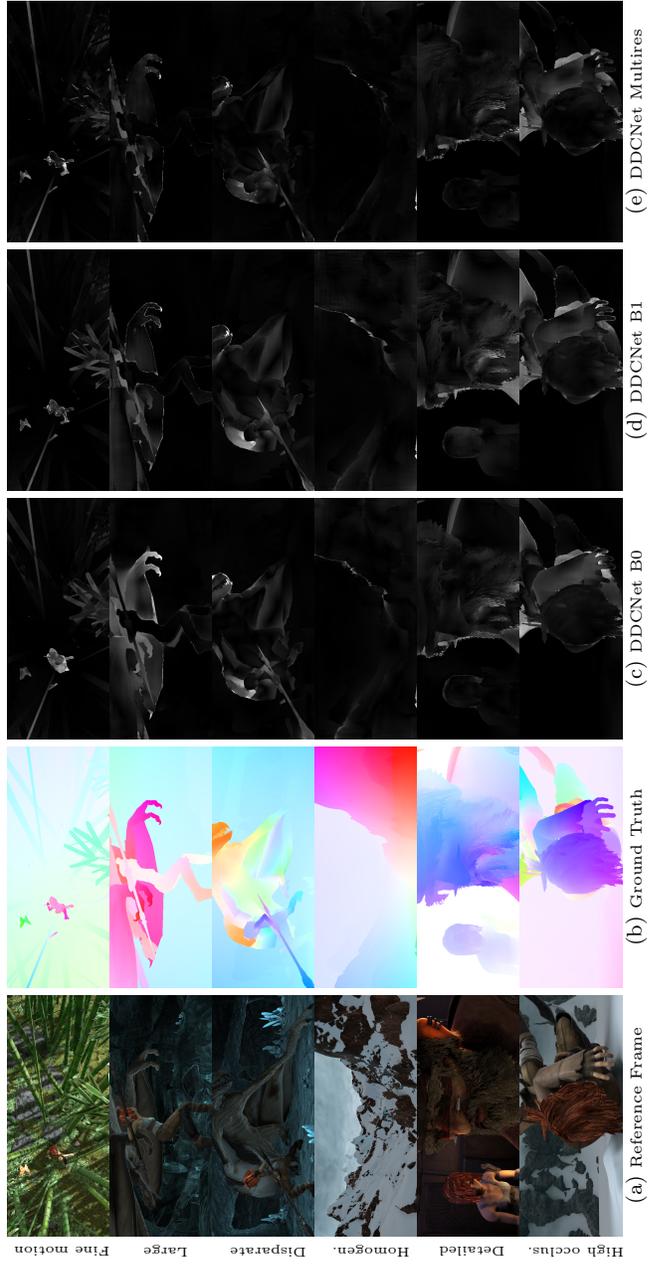

      \centering
      \scalebox{1.0}{\def\scale{0.166}
\def\rootDir{Figures/Qualitative/DDCNets}
\newcommand{\rott}[1]{\fontsize{1}{1}\selectfont{\rotatebox[origin=c]{90}{#1}}}

\newcommand{\IMG}[1]{\includegraphics[align=c,width=\scale\textwidth]{\rootDir/#1}}

\newcommand{\rowImg}[2]{
\xdef\id{#2}
\begin{minipage}[b]{0.01\linewidth}\centering \rott{#1} \end{minipage} & \IMG{GroundTruth/\id-f.png} & \IMG{GroundTruth/\id-gt.png} & \IMG{B0/\id-dis.png} & \IMG{B1/\id-dis.png} & \IMG{Multires/\id-dis.png}
}

\newcommand{\capNew}[1]{\begin{minipage}[b]{\scale\linewidth}\centering\subcaption{\tiny #1}\end{minipage}}
\setlength\tabcolsep{1.5pt}
\begin{tabular}{cccccc}
    \rowImg{Fine motion}{152}\\
    \rowImg{Large}{243}\\
    \rowImg{Disparate}{270}\\
    \rowImg{Homogen.}{364}\\
    \rowImg{Detailed}{391}\\
    \rowImg{High occlus.}{64}\\
    & \capNew{Reference Frame} & \capNew{Ground Truth} & \capNew{DDCNet B0} & \capNew{DDCNet B1} & \capNew{DDCNet Multires} 
\end{tabular}
}
      \decoRule
      \caption[Error maps for DDCNet methods on selected examples]{Optical flow estimation error maps of DDCNet-B0, B1 vs Multires for the optical flow estimates shown in \ref{fig:QualitativeB0vsB1vsMultires}.  Brighter locations indicate locations with larger estimation error.}%
      \label{fig:QualitativeB0vsB1vsMultiresError}%
    \end{sidewaysfigure}

    Figure~\ref{fig:QualitativeB0vsB1vsMultiresError} shows the corresponding error maps for each of the example sequences. The error map represents the magnitude of the difference between the estimated and ground-truth flow vector at each pixel.  Therefore, locations with larger estimation errors will appear brighter and locations with lower estimation errors will appear darker in the error maps.
    
    For the \emph{fine motion} sequence, the Multires model captured the most finer flow profiles and B0 ignored finer motion details and motion boundaries.  Difficulties of the B0 model in detecting larger motions and clear motion boundaries in the \emph{larger motion} sequence (e.g. near the dragon's legs in the scene) is likely due to the smaller extent of its ERF. With a broader ERF extent, and peaking ERF near each reference pixel, the Multires model was able to detect larger motions and with significantly clearer motion boundaries.  In images sequences with multi-directional flows in the scene (evident from multiple colors in the color-coded optical flow) as in the \emph{disparate motion} sequence, we observed that activations of neurons in the latest layers due to multi-directional flows could cancel each other leading to less accurate flow estimates.  Multires was again the best performing model followed by  B0 and B1.  All DDCNet models performed well on \emph{homogeneous texture} sequence with Multires leading the list.  For the \emph{high texture} sequence all of the methods have difficulties identifying fine-grained flow estimates and motion boundaries in the high-textured zones.  Multires and B0 were able to estimate finer flow velocities compared to B1.  For the most challenging \emph{high occlusion} sequence, all DDCNet models faced difficulties in the occluded zones or zones with aperture problem.  Multires again performed better than B0 and B1 models.
    
    In summary, Multires outperforms B0 and B1 models in terms of estimation accuracy with a similar processing time as B0. And Multires is also more accurate in estimating large flows, finer flows, and enforcing clearer motion boundaries in the estimated flows.

    \subsubsection{Visual Inspection of Multi-resolution Estimates in DDCNet-Multires}
\begin{figure}[h]
        \centering
        \scalebox{0.8}{\input{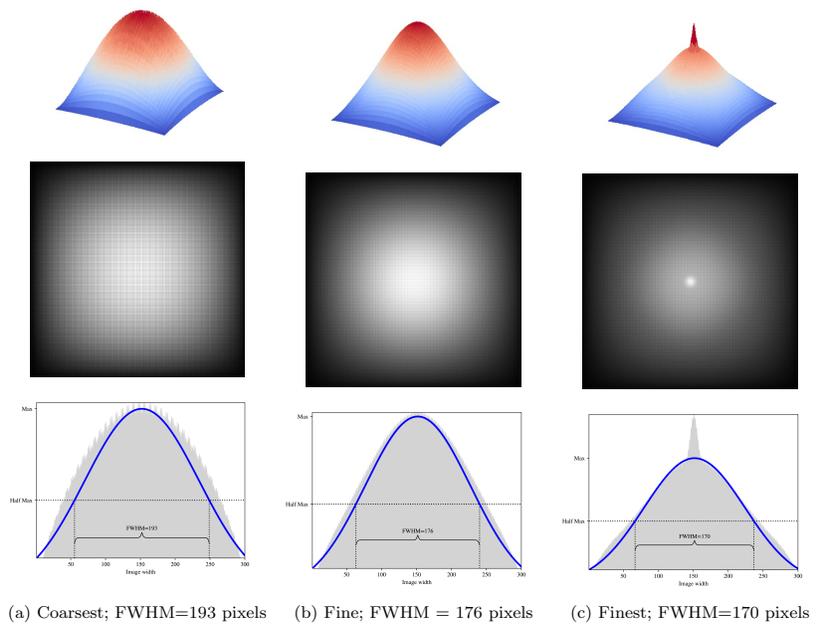}}
        \decoRule
        \caption[ERFs of each of the multi-resolution segments of the DDCNet-Multires]{ERFs of each of the multi-resolution segments of the DDCNet-Multires model corresponding to a) \emph{flow feature extractor}, b) first level \emph{flow feature refiner} segment and c) second-level \emph{flow feature refiner} segment.}
        \label{fig:ErfMultiresCoarseToFine}
    \end{figure}

    Figure~\ref{fig:ErfMultiresCoarseToFine} shows ERFs of each of the multi-resolution segments of the DDCNet-Multires model (one coarse segment and two refiner segments). These ERF maps are obtained by taking derivatives of two intermediate and one final flow layer with respect to the input layer. For example, ERF in Figure~\ref{fig:ErfMultiresCoarseToFine}a is obtained by considering coarse flow layer after \emph{flow feature extractor} sub-net as output of the network (see Figure~\ref{fig:DDCNetMultires}).  For illustrating distinctions of the multi-resolution levels, smaller frame size sequences were used for generating these ERF plots.  
    
\begin{figure}[htbp]
        \centering
        \def\mainLabel{fig:MultiresEstimates}
\begin{tabular}{c}
     \subcaptionbox{Coarsest estimate \label{\mainLabel:a}}{\includegraphics[width=0.44\linewidth]{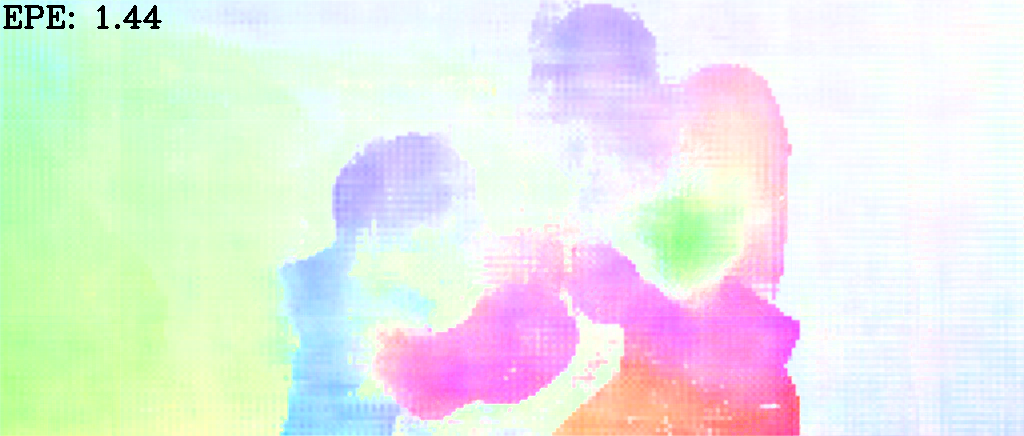} \includegraphics[width=0.44\linewidth]{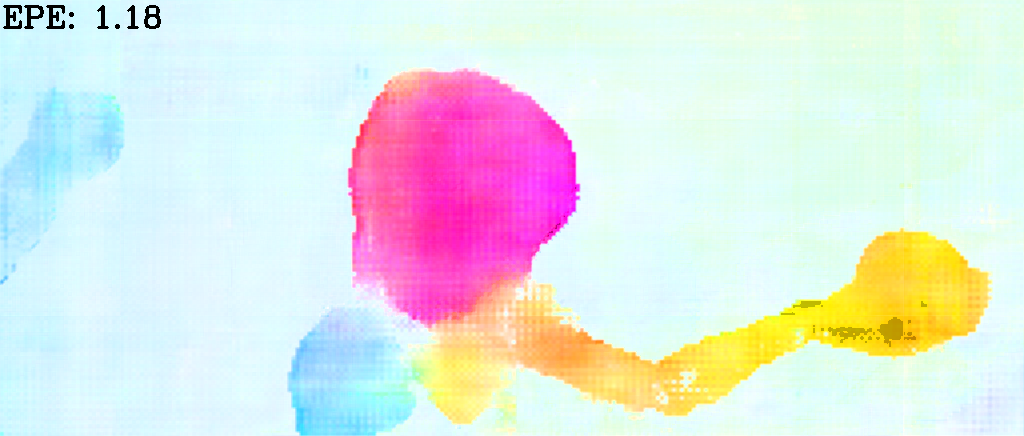}}\\
    \subcaptionbox{Fine estimate \label{\mainLabel:b}}{\includegraphics[width=0.44\linewidth]{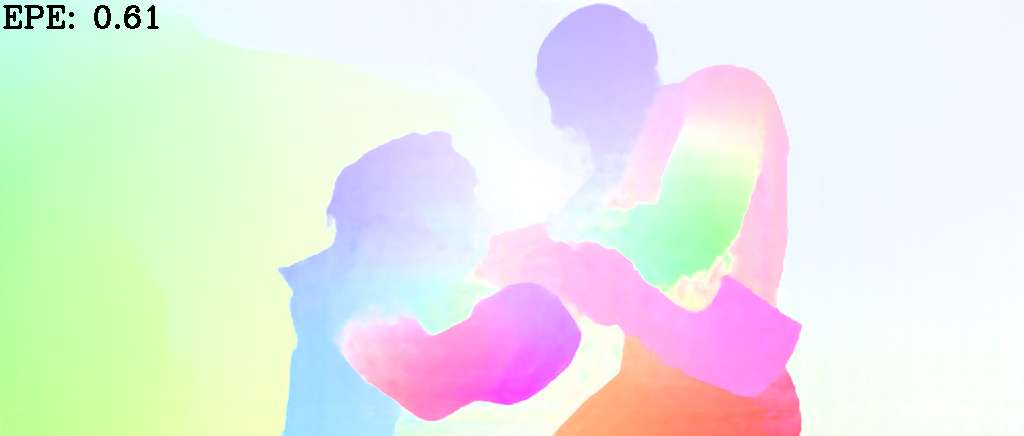} \includegraphics[width=0.44\linewidth]{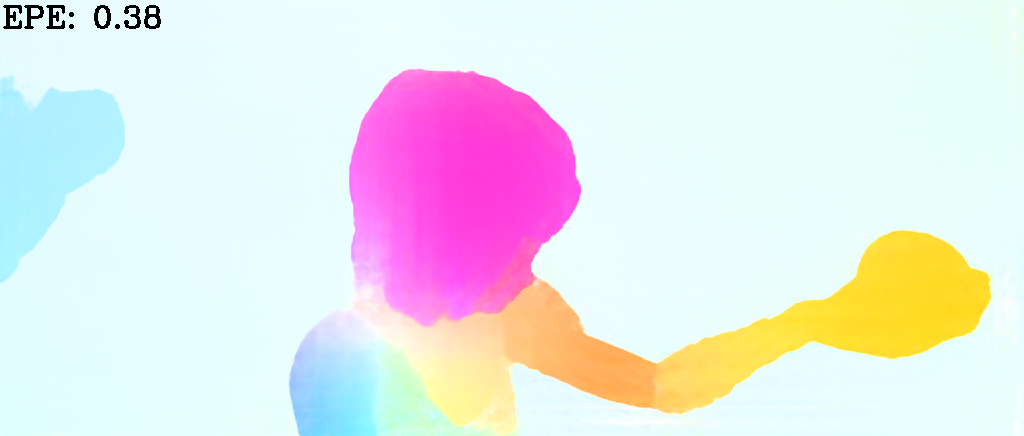}}\\
    \subcaptionbox{Finest estimate \label{\mainLabel:c}}{\includegraphics[width=0.44\linewidth]{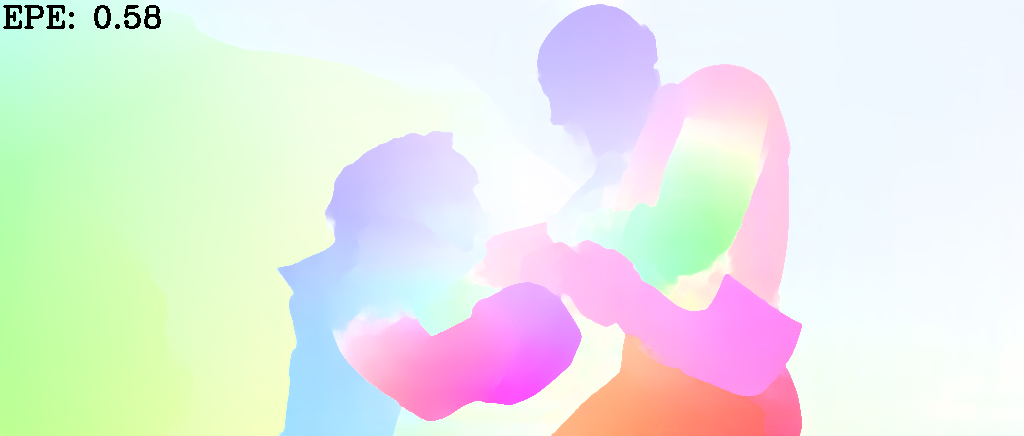} \includegraphics[width=0.44\linewidth]{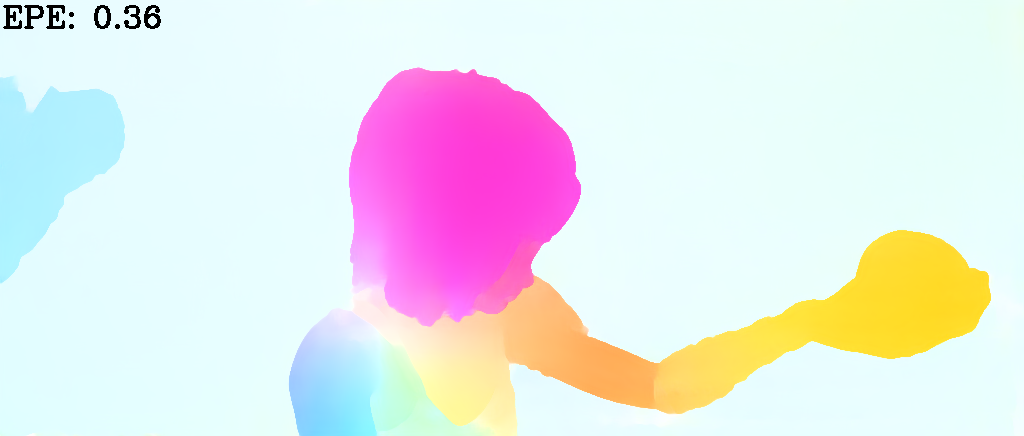}}\\
    \subcaptionbox{Ground truth flow from Sintel \label{\mainLabel:d}}{\includegraphics[width=0.44\linewidth]{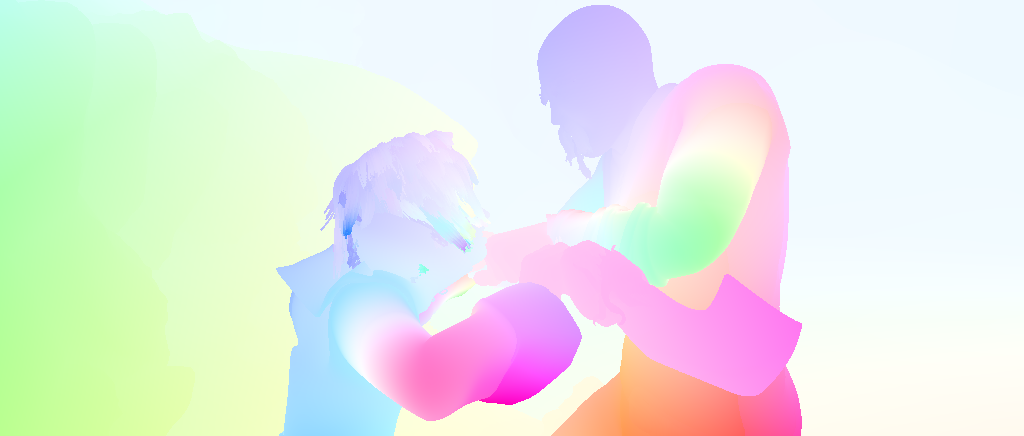} \includegraphics[width=0.44\linewidth]{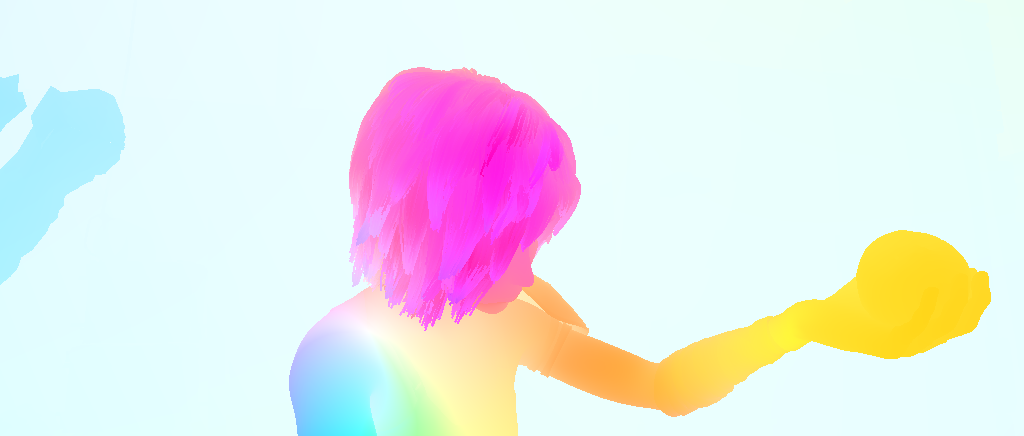}}
\end{tabular}
    
        \decoRule
        \caption[Intermediate and final estimate of DDCNet-Multires]{Intermediate and final flow estimates of DDCNet-Multires: \subref{fig:MultiresEstimates:a} and \subref{fig:MultiresEstimates:b} are intermediate optical flow estimates of the network. \subref{fig:MultiresEstimates:c} is the final estimate from the network.}
        \label{fig:MultiresEstimates}
    \end{figure}

    ERF corresponding to the coarsest output layer is wider with a Gaussian shape but with gridding artifacts. Adding the first flow refiner module minimizes ERF gridding artifacts while retaining a larger ERF extent. The final flow refiner module introduces a sharp peak in the ERF at each of the reference pixel locations emphasizing nearby pixel locations during matching pursuits. This likely helps the network to generate sharper and more accurate results along motion boundaries.
    Figure~\ref{fig:MultiresEstimates} illustrates intermediate optical flow estimates generated within the Multires model at two coarser levels (Figures~\ref{fig:MultiresEstimates} a, b) used to generate a final fine-resolution optical flow estimate (Figure~\ref{fig:MultiresEstimates} c).  The first-level estimate is the coarsest and suffers from gridding artifacts as it is expected based on the ERF of this segment (Figure~\ref{fig:MultiresEstimates} a). Results after the first flow refiner module are smooth and most of the artifacts are resolved (Figure~\ref{fig:MultiresEstimates} b). The final flow refiner module does not improve the accuracy significantly, but it is further refining the estimate and removing even more noises especially along the motion boundaries (Figure~\ref{fig:MultiresEstimates} c).
    
    It is possible to reconfigure the Multires model to generate coarser flow estimates with a faster processing time (suitable for real-time applications) or finer flow estimates with longer processing time. For example, finer flow estimates can be achieved using three levels of feature refiner modules. Such high-resolution finer flow estimates may be especially important when flow estimation is part of a high-level prediction task such as activity recognition or video compression. 
    
\section{Software}
The DDCNet-Multires model along with necessary instructions for running the software are available to the public in the following URL:\\ \href{https://github.com/alisaaalehi/DDCNet}{https://github.com/alisaaalehi/DDCNet}.

\section{Conclusion}
In this work, we have devised a compact deep dilated CNN for dense prediction problems using 1) network design strategies guided by the effective receptive field characteristics of the network and 2) a cascaded sub-net approach to achieve multiresolution capability for handling large heterogeneous motion.  In the cascaded sub-net approach, each sub-net with a varying ERF extent and ERF characteristics provides an optical flow estimate.  Sub-nets with decreasing or varying ERF extents and characteristics are interconnected to achieve a multiresolution capability without image or feature warping.  Thus, DDCNet-Multires avoids the ghosting artifacts and minimizes the vanishing problem.  Desired overall ERF of the network can be achieved by effective combination and arrangement of DDCNet sub-nets with varying ERF shape, extent and smoothness characteristics.  Accuracy of our compact DDCNet-Multires network on standard optical flow benchmark datasets was better than FlowNet-Simple and comparable to LiteFlownet.  In conclusion, our three strategies or recommendations namely 1) preserving spatial information throughout the network, 2) optimal ERF characteristics, and 3) multiresolution through cascaded sub-nets with varying ERF characteristics are useful for building more compact networks for dense prediction problems using standard network elements.

\section*{Acknowledgement}
The research was supported in part by financial support in the form of a Herff Fellowship from the Herff College of Engineering, The University of Memphis (UoM); tuition fees support from the Department of Electrical and Computer Engineering, UoM; a summer student fellowship from the \emph{Fight for Sight} organization; and in part by the National Institutes of Health, National Eye Institute Grant EY020518.

 %
 


\pagebreak
\bibliographystyle{plainnat}
\bibliography{References.bib}
\end{document}